\documentclass[10pt, twocolumn, letterpaper]{article}
\usepackage{cvpr}              % To produce the CAMERA-READY version

\pdfoutput=1
\usepackage{amsmath,amsfonts}
\usepackage{algorithmic}
\usepackage{algorithm}
\usepackage{array}
\usepackage{textcomp}
\usepackage{stfloats}
\usepackage{url}
\usepackage{verbatim}
\usepackage{graphicx}
\usepackage{cite}
\def\BibTeX{{\rm B\kern-.05em{\sc i\kern-.025em b}\kern-.08em T\kern-.1667em\lower.7ex\hbox{E}\kern-.125emX}}
\usepackage{color}
\usepackage{balance}
\usepackage{pifont}
\usepackage{tabularx}
\usepackage{booktabs}
\usepackage{multirow}
\usepackage{etoolbox}
\usepackage{amssymb}
\makeatletter
\patchcmd{\@makecaption}
  {\scshape}
  {}
  {}
  {}
\makeatletter
\patchcmd{\@makecaption}
  {\\}
  {.\ }
  {}
  {}
\makeatother

\usepackage[dvipsnames, table]{xcolor}

\definecolor{cvprblue}{rgb}{0.21,0.49,0.74}
\usepackage[pagebackref,breaklinks,colorlinks,citecolor=cvprblue]{hyperref}

\def\paperID{3832} % *** Enter the Paper ID here
\def\confName{CVPR}
\def\confYear{2024}

\title{Consistency Prototype Module and Motion Compensation for Few-Shot Action Recognition (CLIP-CP$\mathbf{M^2}$C)}

\author{Fei Guo, Li Zhu$^*$, YiKang Wang, Han Qi\\
School of Software Engineering, Xi'an Jiaotong University, China\\
{\tt\small co.fly@stu.xjtu.edu.cn funnyq@stu.xjtu.edu.cn zhuli@xjtu.edu.cn  qihan19@stu.xjtu.edu.cn}
}

\begin{document}
\maketitle
\begin{abstract}
Recently, few-shot action recognition has significantly progressed by learning the feature discriminability and designing suitable comparison methods. 
Still, there are the following restrictions. 
(a) Previous works are mainly based on visual mono-modal. Although some multi-modal works use labels as supplementary to construct prototypes of support videos, they can not use this information for query videos. The labels are not used efficiently. 
(b) Most of the works ignore the motion feature of video, although the motion features are essential for distinguishing. 
We proposed a Consistency Prototype and Motion Compensation Network(CLIP-CP$M^2$C) to address these issues. 
Firstly, we use the CLIP for multi-modal few-shot action recognition with the text-image comparison for domain adaption.
Secondly, in order to make the amount of information between the prototype and the query more similar, we propose a novel method to compensate for the text(prompt) information of query videos when text(prompt) does not exist, which depends on a Consistency Loss. 
Thirdly, we use the differential features of the adjacent frames in two directions as the motion features, which explicitly embeds the network with motion dynamics. We also apply the Consistency Loss to the motion features.
Extensive experiments on standard benchmark datasets demonstrate that the proposed method can compete with state-of-the-art results. Our code is available at the URL:
\url{https://github.com/xxx/xxx.git}.
\end{abstract}    
\section{Introduction}
\label{sec:intro}

In the realm of computer vision, human action recognition has been widely used in various fields. 
Video action recognition is an important branch of human action recognition. 
The application of deep learning in this field has greatly advanced its progress. 
To achieve this success, we need many manually annotated videos. 
Unfortunately, acquiring such labeled data in production environments is both time-consuming and labor-intensive, rendering it impractical.
A promising solution to this challenge is few-shot action recognition, a paradigm studied by researchers in recent years. 
This approach enables action recognition without many expensive labels.
Few-shot action recognition can be classified according to the following two criteria. 
The first criterion pertains to "How to perform the calculation?" and includes three categories:
(1) memory-based methods \cite{zhu2018compound,zhang2020few1,qi2020few-smfn}. 
(2) data-enhancement-based methods \cite{kumar2019protogan,fu2020depth}. 
(3) metric-based comparison methods
\cite{cao2020few-otam, bishay2019tarn, perrett2021temporal-trx, thatipelli2022spatio-strm, zhang2020few-ARN,tan2019learning-FAN, lu2021few-cmot, li2022hierarchical-HCR}. 
Metric-based comparison, a crucial method, further divides into a comparison of global semantics and a comparison of temporal information between video frames.
The second criterion considers "What kind of content can be used for calculation?" and encompasses three categories:
(1) visual mono-modal 
\cite{cao2020few-otam, bishay2019tarn, perrett2021temporal-trx, thatipelli2022spatio-strm, tan2019learning-FAN, wu2022motion, wang2023molo}
(2) visual multi-modal \cite{fu2020depth, zhu2020label, wang2022task, wanyan2023active}. 
(3) visual-text multi-modal \cite{zhang2020few1, su2021vdarn, wang2021semantic,ni2022multi-modal-CLIP, wang2023CLIP}. 
Our work belongs to the category of metric-based methods and also belongs to the category of visual-text multi-modal.

Through the analysis of previous work, it becomes apparent that in mono-modal approaches, extracting more accurate statistical characteristics of video distribution poses a challenge due to the limited samples in each episode. In contrast, multi-modal methods leverage complementary information, providing a solution to the issue caused by insufficient samples. 
In recent years, CLIP \cite{CLIP-origin2021}, a large-scale visual-language pre-processing model, has used massive image-text pairs from the Internet as training samples for comparative learning, thereby obtaining robust image-text matching accuracy. Researchers have successfully applied the CLIP model in various downstream tasks \cite{d-CLIP-guzhov2022audioCLIP, d-CLIP-li2022grounded-detect, d-CLIP-xu2022groupvit, d-CLIP-zhang2022can-depth, wang2021actionCLIP}, demonstrating its effectiveness through domain adaptation. In particular, in the domain of action recognition, researchers have conducted extensive studies on the migration of CLIP.

The motivation of CLIP in few-shot action recognition is based on that pre-trained visual-language models can be generalized to video tasks. The visual encoder is introduced for the visual features in MORN(2022) \cite{ni2022multi-modal-CLIP}. After flattening, text prototypes are obtained. Using the prototype-enhanced module, the visual and text prototypes are operated with the weighted sum for multi-modal prototypes. Its comparison method is similar to the TRX(2021) \cite{perrett2021temporal-trx}. There is no Domain Adaption for this work. CLIP-FSAR(2023) \cite{wang2023CLIP} also belongs to the multi-modal based on CLIP. This work uses a handcrafted template to create a prompt and then average the frames of each video. The comparison between the text feature and the averaged frame feature acts as the Domain Adaption
%(the idea is similar to A5, CLIP4CLIP, XCLIP and %ActionCLIP \cite{wang2021actionCLIP, ni2022expanding-%XCLIP, luo2022CLIP4CLIP, ju2022prompting-A5}). 
The text and frame features are concatenated in this work and sent to a Transformer to get the multi-modal prototype.

There are three problems with these CLIP-related works: 
The first is that the existing works use the label (prompt) information and the visual information to create the prototype. However, they can not acquire the label information for the query representation. So, the amount of information between the support prototype and the query representation is not equal. In the paradigm of few-shot learning, it can be seen that we can acquire the labels of the query video in the meta-training. How to find a suitable method to effectively utilize the query labels in the meta-training and extend this method to the meta-testing without query labels needs to be studied. 
The second is the operation of the existing works for the prototype is simple, and the temporal information perhaps cannot be fully excavated. For example, MORN \cite{ni2022multi-modal-CLIP} uses a weighted sum for visual and text, CLIP-FSAR\cite{wang2023CLIP} considers a simple concatenation for text and frames. 
The third is that the prototype and the query representation for metric comparison are usually generated from the normal frames. It ignores the motion feature, which can compensate for the information the normal frames miss. From now on, no one has introduced the motion features to CLIP-based few-shot action recognition.

To cope with these problems, we propose a model named Consistency Prototype and Motion Compensation based on the CLIP. 
\textbf{The first part} is Consistency Prototype Module with Consistency Loss. 
We introduce a Feature Enhancement with a Transformer at its core, utilizing a token and frames feature as input for prototype construction. During meta-training, the prompt embedding serves as the token with both support and query.
Meanwhile, a random vector also acts as the token with support and query. 
The Feature Enhancement processes all support and query samples with different tokens. 
The objective is to ensure the representations of support or query with different tokens become close to each other through a Consistency Loss, and the random vector can replace the prompt embedding effectively. 
In meta-testing, we concatenate the real prompt embedding in front of the support and use a random vector as a fake prompt embedding in front of the query for comparison.
\textbf{The second part} is Motion Compensation. In this part,  
we employ a CNN network to handle the features of each frame and perform a difference operation between the features before the CNN operation and their adjacent features obtained through CNN. This operation is carried out in two directions, and the resulting motion features are obtained, which can serve as compensation for the normal frames.

In summary, our contributions are as follows:
%\begin{itemize}
(1) We use the CLIP model as the backbone for the few-shot action recognition and migrate the image-text matching model to the task of video-text matching through a Domain Adaption module which is similar to CLIP-FSAR.
(2) We propose a Consistency Prototype Module based on visual features from the visual encoder and the prompt embedding from the text encoder. In this module, we proposed a new paradigm to generate text-video representation using a fake prompt vector for query video during the comparison. To the best of our knowledge, it is the first work to use a fake prompt embedding in generating query representation for few-shot action recognition based on multi-modal.
(3) We introduce a Motion Compensation mechanism to calculate adjacent frames and obtain dynamic features of the video. The method for constructing the motion prototype mirrors the process for normal frame features. This marks the first attempt to incorporate motion features into CLIP-based few-shot action recognition.
(4) On the five benchmarks as HMDB51, UCF101, Kinetics, and Something-Something-V2-Full(Small), extensive experiments demonstrate that our CLIP-CP$\mathrm{M^2}$C has a favorable performance and can compare with the state-of-the-art methods.
%\end{itemize}

\section{Related Work}
\label{sec: related work}

\subsection{CLIP modal}
The CLIP model \cite{CLIP-origin2021} is a pre-trained neural network model released by OpenAI in early 2021 for matching images and texts. It has been a classic in multi-modal research in recent years. The model directly uses a large amount of text-image pairs from the Internet for pre-training and has achieved the best performance in many tasks. Using these pairs for training can not only solve the high-cost problem of obtaining labeled data but also make it easier to obtain a model with solid generalization ability depending on the amount of data from the network, which is relatively large, and the data is pretty different. The idea of the CLIP model is to improve the model's performance by learning the matching relationship between image and text. Precisely, the CLIP model consists of two main branches: an encoder with CNN-RN \cite{he2016deep-resnet} or the VIT \cite{vision-transformer} for processing images and a Transformer model \cite{vaswani2017attention} for processing text. Both branches are trained to map the input feature into the same embedding space and force image-text pairs to be close in the embedding space. Except for the image-text, some other tasks also use the CLIP model for the multi-modal comparison, and how to give a perfect method for the Domain Adaption is essential.

\subsection{Few-shot Image Classification}
Metric-based, Model-based, and Data-augmentation-based methods are the three main categories for few-shot image classification.
\textbf{Metric-based method}: 
These methods aim to learn feature embedding under a certain distance computing and have good generalization ability. The samples in novel categories can be classified accurately using different distance classifiers. 
Such as cosine similarity \cite{vinyals2016matching—few_shot_learning5, ye2020few-embedding-adaptation-few_shot_learning8}, 
Euclidean distance \cite{pahde2021multimodal-few-augmentation4, yoon2019tapnet}, etc. 
\cite{simon2020adaptive-few_shot_learning1,snell2017prototypical_few_shot_learning2,sung2018learning-relation-networkfew_shot_learning3,melekhov2016-siamese—few_shot_learning4,vinyals2016matching—few_shot_learning5,liu2018learning-TNP-few_shot_learning7, ye2020few-embedding-adaptation-few_shot_learning8} belong to this category.
\textbf{Model-based methods}:
These methods aim to quickly update parameters on few-shot samples through the design of the model structure, directly establishing a mapping function between the input $x$ and the predicted value $p$. \cite{finn2017model-MAML-few-shot-model1, rusu2018-metaL-few-shot-model3, ravi2017optimizationL-few-shot-model4, 9999670-generalization-of-MAML}  belong to this category.
\textbf{Data-Augmentation-based methods}:
These methods use data generation strategies to improve the generalization ability of few-shot models. \cite{chen1804semantic-few-augmentation3, ratner2017learning-few-augmentation2, perez2017effectiveness-few-augmentation1,pahde2021multimodal-few-augmentation4,10035001-generation-few-shot-learning} follow this category.

\subsection{Few-shot action recognition}
Many methods of few-shot action recognition have emerged recently because they can solve the problem that getting the video labels is labor-expensive and time-consuming. We can divide FSAR into the mono-modal and multi-modal. 

\textbf{Mono-modal.}
Most works in the field of few-shot action recognition follow this paradigm, and we can summarize this kind of work as follows:
CMN\cite{zhu2018compound} encodes the video representation using multi-significant embedding, and the key and value are stored in memory and can be updated easily. 
SMFN\cite{qi2020few-smfn} proposes to solve the few-shot video classification problem by learning a set of SlowFast networks enhanced by memory units. 
ProtoGAN\cite{kumar2019protogan} uses a generative adversarial network with specific semantics as the condition to synthesize additional samples of novel classes for few-shot learning. 
OTAM \cite{cao2020few-otam} computes the distance matrix following the DTW\cite{muller2007dynamic}method and makes a strict alignment. 
TARN\cite{bishay2019tarn} first introduced the attention mechanism for temporal alignment, and its alignment is frame level. 
TRX\cite{perrett2021temporal-trx} uses the Cross-Transformer to deal with frame tuples, then gets the tuples' feature embedding alignment. 
STRM \cite{thatipelli2022spatio-strm} is the improvement of the TRX and adds some pre-processing for feature enrichment. 
ARN \cite{zhang2020few-ARN} focuses on robust similarity, spatial and temporal modeling of short-term and long-term range action patterns via permutation-invariant pooling and attention. 
CMOT\cite{lu2021few-cmot} uses the Optimal Transport theory to compare the content of the video, not only focusing on the ordered alignment but also focusing on the video semantics.
HyRSM\cite{wang2022-hybrid} learns task-specific embeddings by utilizing hybrid relation modeling.
MTFAN \cite{wu2022motion} leverages the motion patterns extracted from videos and improves the transferability of feature embedding.

\textbf{Multi-modal.}
There are several works belonging to multi-modal for few-shot action recognition.
CMN-J\cite{zhu2020label} belongs to this category. On the one hand, it introduces a label-independent repository to store correlations between unlabeled data and target examples. On the other hand, it uses multi-modal features that contain RGB and optical flow features. ARCMN\cite{zhang2020few1} introduces the cross-modal: one modal is the video content, and another modal is the action label.
AMeFu-Net\cite{fu2020depth} contains the RGB feature and depth features.
MORN(2022)\cite{ni2022multi-modal-CLIP} introduces CLIP to few-shot action recognition and brings multi-modal. It uses the semantics of labels to enhance the prototype and uses TRX for metric comparison.
CLIP-FSAR\cite{wang2023CLIP} is a work also related to CLIP. Firstly, it uses the text-visual comparison for Domain Adaption. Then, each episode uses prompt embedding to enhance the visual feature and construct a multi-modal prototype.
\section{Methodology}
\label{sec: method}

\subsection{Problem Setting}
In the field of few-shot action recognition, the video datasets are split into $D_{train}$, $D_ {test}$, $D_ {val}$, all the split datasets should be disjoint, which means there are no overlapping classes between each split dataset. In the $D_{train}$, it contains abundant labeled data for each action class, while there are only a few labeled samples in the $D_ {test}$, and the $D_ {val}$ is used for model evaluation during the training episode. No matter $D_{train}$, $D_ {test}$, or $D_ {val}$, they all follow a standard episode rule. Episode, also called task, occurs during the training, testing, or validation. In each episode, $N$ classes with $K$ samples in $D_{train}$, $D_ {test}$, or $D_ {val}$ are sampled as ``support set". The samples from the rest videos of each split DB are sampled as ``query set", just as $P$ samples are selected from $N$ classes to construct the ``query set". 
The goal of few-shot action recognition is to train a model using $D_{train}$, which can be generalized well to the novel classes in the $D_ {test}$ only using $N\times K$ samples in the support set $D_ {test}$. Let $q = ({q_1, q_2, \cdots , q_m })$ represents a query video with $m$ uniformly sampled frames. We use $C$ to represent the set of classes, $C =\{c_1,\cdots, c_N\}$, and the goal is to classify a query video $q$ into one of the classes $c_i \in C$. In our work, the support set is defined as $S$, and the query set is defined as $Q$. For the class $c$, the support set $S_c$  can be expressed as $S_c = \{s_c^1, \cdots ,s_c^k \cdots, s_c^K\}, 1 \leq k \leq K$, and $s_c^k= ({s_c^{k:1}, \cdots, s_c^{k:m}}),  1 \leq k \leq K$, $m$ is the frame number.

\begin{figure*}
\setlength{\belowcaptionskip}{-0.5cm} %调整图片标题与下文距离
\centering
\includegraphics[width=18cm]{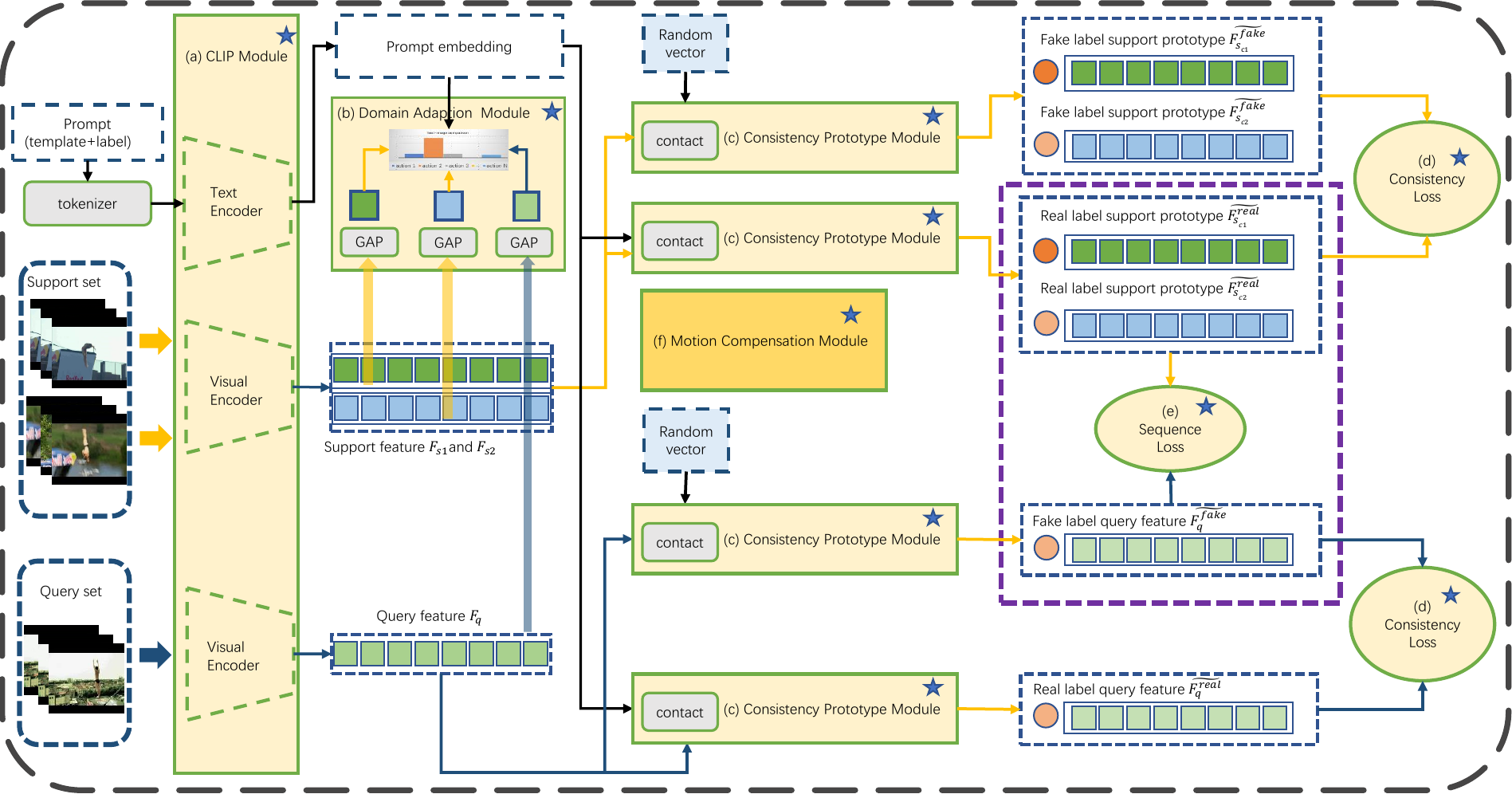}
\caption{\label{fig:CLIP-few-shot-framework}The Framework of CLIP-CP$\mathrm{M^2}$C. In this illustration, the workflow of our model can be seen as follows: 
\textbf{(a)} Pre-trained CLIP Model(Visual encoder and Text encoder) . 
\textbf{(b)} Domain Adaption Module(DAM). 
\textbf{(c)} Consistency Prototype Module(CPM).
\textbf{(d)} Consistency Loss(CL).
\textbf{(e)} Sequence Distance.
\textbf{(f)} Motion Compensation Module(MC).
The green solid line regions filled with yellow represent the modules and the loss, and the blue dashed line regions represent the data or the features in each step. 
The purple dashed line region represents the metric comparison. 
Also, there are some other simple operations such as as:``tokenizer", ``GAP" and ``contact".
The black arrows represent semantic data flow, the yellow arrows represent support data flow, and the blue arrows represent query data flow. 
For simplicity, we use a 2-way 1-shot setting in the illustration. We use blue stars to mark the (a) to (f).
} 
\label{fig:1}
\end{figure*}

\subsection{Overall framework}
We propose a framework for few-shot action recognition as \cref{fig:1}. There are 6 parts of our model: 
(a) CLIP Module. It uses the visual encoder to encode video frames and the text encoder to encode the label prompt.
(b) Domain Adaption Module. This module uses the average aggregation for each video's frames and compares prompt embedding and averaged frame embedding. An across-entropy loss is used to force the adaption.
(c) Consistency Prototype Module. A Transformer is used to deal with the feature concatenated by the frames and the video's label prompt embedding (real or fake). Then, the features will focus on not only the frame information but also the global semantic information.
(d) Consistency Loss. It is the L2 Distance between the real-labeled feature and the fake-labeled feature. The first one is generated by the concatenation of real prompt embedding and visual embedding, and the second one is generated by the concatenation of a random vector and visual embedding.
(e) Sequence Loss. It depends on the distance between the features of support frames and query frames.
(f) Motion Compensation Module. In this module,  the feature difference between adjacent video frames is calculated as the motion feature for compensating the normal frame feature.

\subsection{Consistency Prototype Module with Consistency Loss (CPM)} \label{Consistency_Prototype_Module}

Currently, some work \cite{ni2022multi-modal-CLIP}, \cite{wang2023CLIP} focus on the multi-modal based on the CLIP for few-shot action recognition. Besides using CLIP as the backbone, they all pay attention to the prototype and the comparison between the prototype and the query representation. All these works only use the prompt information to construct the prototype with visual information. 
In meta-testing, there is no label for the query beforehand, so the query only uses visual information for comparison. 
Undoubtedly, this method leads to inconsistent information content between the prototype and the query. \textbf{If we can generate a fake prompt embedding that can instead of the real prompt embedding even though the real prompt information does not exist in the meta-testing stage?}
Our Consistency Prototype Module can solve this problem.
The innovation is based on the paradigm of meta-learning, and the fact is that we know the label of query in the meta-training stage.
Our idea is to use a random vector whose dimension is the same as the real prompt embedding to compensate for the missing real prompt embedding. 

\subsubsection{Feature Enhancement} \label{subsubsec: feature enhancement}
\begin{figure}
\setlength{\belowcaptionskip}{-0.5cm} %调整图片标题与下文距离
\centering
\includegraphics[width=8cm]{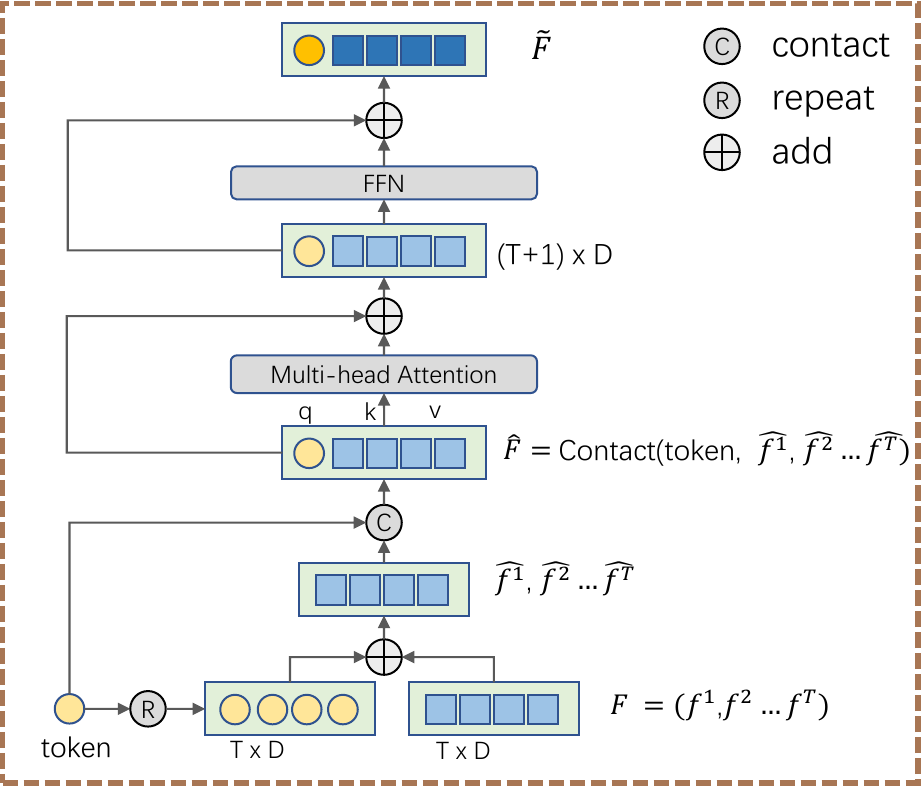}
\caption{\label{fig:transformer-prototype}
The illustration shows the structure of Feature Enhancement. The core is a Transformer and the $token$ could be a prompt embedding or a random vector.
 \small{\textcircled{\scriptsize{C}}}\small means contact operation,
 \small{\textcircled{\scriptsize{R}}}\small means repeat operation.
 $\oplus$ means sum operation.
} 
\label{fig:2}
\end{figure}

In action recognition, some works focus on adapters for the CLIP. But for few-shot action recognition, an enhanced prototype is also very important. 
Given the features $F = [f^1, f^2, ..., f^T]$ from the visual encoder and the $token$ (prompt embedding) from the text encoder, \cref{eq1} to \cref{eq3} according to \cref{fig:2} shows the workflow of Feature Enhancement.

\begin{small}
\begin{equation}  \label{eq1}
   \setlength\abovedisplayskip{-2pt}
   \setlength\belowdisplayskip{-2pt}
   [\hat{f^1}, \hat{f^2}, ..., \hat{f^T}] = Repeat(token) + [f^1, f^2, ..., f^T] 
\end{equation}
\end{small}

\begin{small}
\begin{equation} \label{eq2}
   \setlength\abovedisplayskip{-2pt}
   \setlength\belowdisplayskip{-2pt}
    \hat{F}= Contact(token, \hat{f^1}, \hat{f^2}, ..., \hat{f^T})
\end{equation}
\end{small}

\begin{small}
\begin{equation} \label{eq3}
   \setlength\abovedisplayskip{-2pt}
   \setlength\belowdisplayskip{-2pt}
    \widetilde{F} = \mathbb{T}(\hat{F} + f_{pos}) \\
\end{equation}
\end{small}
where $token\in R^D$, $D$ is the dimension of prompt embedding and each frame $f^t$.
The $Repeat()$ copies $token$ to T copies,
so $Repeat(token)  \in R^ {T \times D}$.
After the $contact$, $\hat{F} \in R^{(T+1) \times D}$.
$f_{pos} \in R^{(T+1) \times D}$ means the position embeddings to encode the position.
$\mathbb{T}$ is the Transformer operation that contains the Multi-head Attention and FFN, 
so $\widetilde{F} \in R^{(T+1) \times D}$.

\subsubsection{Consistency Loss}
In the meta-training,
we assume $F_{s_c^i} = [f_{s_c^i}^1, f_{s_c^i}^2, ..., f_{s_c^i}^T]$ is the feature of $ith$ support sample under class $c$, and 
$F_{q_p^i} = [f_{q_p^i}^1, f_{q_p^i}^2, ..., f_{q_p^i}^T]$
is the feature of $ith$ query sample under class $p$. 
We use the text encoder to encode the prompt of support class $c$ and query class $p$, and get the prompt embedding as $token_c$ and $token_p$
Then, we send the $token_c$ and $F_{s_c^i}$ to the Feature Enhancement Module \cref{subsubsec: feature enhancement} for temporal and semantic information aggregation. The same operation is for $token_p$ and $F_{q_p^i}$.
Then we get the output $\widetilde{F_{s_c^i}^{real}}$ as real-labeled support features and $\widetilde{F_{q_p^i}^{real}}$ as real-labeled query features. 
Meanwhile, we create a $token_r$, a Gauss random vector with $0$ means and $1$ variance. We send $token_r$ and $F_{s_c^i}$ to the Feature Enhancement Module for temporal and semantic aggregation. The same operation is for another $token_r$ and $F_{q_p^i}$. Also, we get the output $\widetilde{F_{s_c^i}^{fake}}$ as fake-labeled support features and $\widetilde{F_{q_p^i}^{fake}}$ as fake-labeled query features. 

Through the Feature Enhancement Module, if the random vector can replace the real prompt embedding, ${F_{s_c^i}^{fake}}$ should be close to ${F_{s_c^i}^{real}}$, and also ${F_{q_p^i}^{fake}}$ should be close to ${F_{q_p^i}^{real}}$. So, we define a Consistency Loss $L_{con}$.
\begin{small}
\begin{equation} \label{eq4}
   \setlength\belowdisplayskip{-0.5pt}
    L_{con} = \sum_{p=1}^{N}\sum_{i=1}^{P}||\widetilde{F_{q_p^i}^{fake}} - \widetilde{F_{q_p^i}^{real}} ||^2  + \sum_{c=1}^{N}\sum_{i=1}^{K}||\widetilde{F_{s_c^i}^{fake}} - \widetilde{F_{s_c^i}^{real}}||^2
\end{equation}
\end{small}
where the \cref{eq4} describes the setting of N-Way K-Shot P-query in the meta-training stage of each episode.

The $L_{con}$ will be sent to the Backpropagate, forcing the $L_{con}$ to be small. The random vector could replace the real prompt embedding when the $L_{con}$ is small enough.

\subsubsection{Consistency Prototype and query feature}
In our work, we assume that the distribution of samples in the meta-training stage and the meta-testing stage are similar. So, we think the random vector that can replace the real prompt embedding in the meta-training stage can also be used in the meta-testing stage. 
We use $\widetilde{F_{s_c^i}^{real}}$ with the Mean pooling operation to create the prototype as follows:
\begin{small}
\begin{equation} \label{eq5}
    \widetilde{F_{s_c}} = \frac{1}{K}\sum_{i=1}^{K}\widetilde{F_{s_c^i}^{real}}
\end{equation}
\end{small}
where $K$ is the number of $shot$ under the category $c$ and $\widetilde{F_{s_c}}$ is the prototype of category $c$.

Although we can not be concerned about the real prompt and the real category of the query during comparison, 
we could use a random $token_r$ and $F_q = [f_q^1, f_q^2, ..., f_q^T]$ to generate the query feature $\widetilde{F_{q}^{fake}}$ in both the meta-training stage and the meta-testing stage for comparison with the prototype. 
In \cref{fig:1}, the purple dashed region represents the comparison between the support prototype and query. 

\subsection{Motion Compensation (MC)}
We get inspiration from the MoLo \cite{wang2023molo} and MTFAN \cite{wu2022motion} and try to introduce motion features in our work. So far, no works have used motion features that are distinguishable in action recognition based on the CLIP few-shot model. We consider the motion feature can also be fused with the semantic feature.
Some few-shot action recognition works use 3DCNN to extract motion features, while others use optical flow to represent motion features. It is expensive for these methods. We have a Motion Compensation Module that uses the difference between adjacent frames to represent the motion features. Specifically, we perform forward difference and backward difference on the features of the video frames encoded by the visual encoder.
See the \cref{eq6}.
\begin{small}
\begin{equation}  \label{eq6}
 \begin{split}
    &m^t_b = f^{t}_x - \Phi(f^{t+1}_x)\\
    &m^t_f = f^{t+1}_x  -  \Phi(f^t_x) 
\end{split}
\end{equation}
\end{small}
where $\Phi$ is a sequence of operations that contains several convolution, batch normalization, and ReLU, $m^t_b$ is a backward motion feature, and $m^t_f$ is a forward motion feature, $t \in \{1,..., T-1\}$, $x$ represents a support or query.

We average the forward and backward motion features respectively, and then can get the global motion representation of two directions. See the \cref{eq7}.
\begin{small}
\begin{equation}  \label{eq7}
 \begin{split}
   \setlength\abovedisplayskip{-2pt}
   \setlength\belowdisplayskip{-2pt}
    &m^g_b = \frac{1}{T-1}\sum_{t=1}^{T-1}GAP(m^t_b)\\
    &m^g_f = \frac{1}{T-1}\sum_{t=1}^{T-1}GAP(m^t_f)
\end{split}
\end{equation}
\end{small}

 After that, we add the global motion representation (forward/backward) and the motion features at each timestamp (forward/backward). In the end, we aggregate the forward and backward features for each video as the final motion features. See the \cref{eq8} and \cref{eq9}
\begin{small}
\begin{equation}  \label{eq8}
\setlength\belowdisplayskip{-2pt}
 \begin{split}
     &m^t_b =  m^t_b  + m^g_b \\
     &m^t_f =  m^t_f  + m^g_f
\end{split}
\end{equation}
\end{small}

\begin{small}
\begin{equation}  \label{eq9}
 \begin{split}
     &m^t =\frac{1}{2}(m^t_b  + m^t_f)
\end{split}
\end{equation}
\end{small}
where $t \in \{1,...T-1\}$.

Then, we send the $[m^1,...m^{T-1}]$ with the corresponding prompt embedding or a random vector to the Consistency Prototype Module and get the $\widetilde{F^{'}}$. 
In the meta-training stage, Consistency Loss for motion is similar to \cref{eq4}.

\subsection{Domain Adaption Module (DAM)} \label{dam}
We adopt the CLIP model, which inherits the excellent accuracy of image-text matching and reduces the demand for computing resources in training. But still, we need a Domain Adaptation to adapt our video-text matching task to the image-text matching task of CLIP. For the text branch, through the text encoder, we can acquire the prompt embedding represented as $ \{token_i\}, i \in \{1, C\}$, $C$ is the number of categories in meta-training. Following the method of Clip4Clip \cite{luo2022CLIP4CLIP}, and ActionClip \cite{wang2021actionCLIP}, we use the Mean Polling to get the video representation after the visual encoder, which encodes all the frames of each video. Then, we use the cosine similarity to get the similarity between the video representation and text feature. 
In the meta-training stage, under the $N-way$, $K-shot$, $P-query$ setting in each episode, through the visual encoder, the support feature set is $F_s= \{F_{s_1}, F_{s_2}...F_{s_{NK}}\}$ and the query feature set is $F_q= \{F_{q_1}, F_{q_2}...F_{q_{NP}}\}$, so the total number of videos for the cosine similarity is $NK+NP$.
\begin{small}
\begin{equation} \label{eq10}
p^{text-video}(y=i|v) = \frac{exp(sim(GAP(F_v), w_i)/t)}{ \sum_{j=1}^{C}{exp(sim(GAP(F_v), w_j)/t)}}
\end{equation}
\end{small}
where $F_v$ is a video representation in a certain episode, it can belong to the $F_s$ or $F_q$, and the $t$ denotes a temperature factor. 
To adapt the difference between domains, the matching loss between text and video is used to force the alignment of the two pieces of information.

\subsection{Comparison and Loss} \label{section: comparison}
\textbf{Distance.}
There are two kinds of sequence distance. One is for the frame level between query and support prototype, and the other is for the motion level. See \cref{eq11} and \eqref{eq12}.
\begin{small}
\begin{equation} \label{eq11}
   \setlength\belowdisplayskip{-1pt}
d(\widetilde{F_{s_c}}, \widetilde{F_q^{fake})} = Distance([\widetilde{f_{s_c}^1}...\widetilde{f_{s_c}^T}],[\widetilde{f_q^1}...\widetilde{f_q^T}])
\end{equation}
\end{small}
\begin{small}
\begin{equation}  \label{eq12}
d(\widetilde{F_{s_c}^{'}}, \widetilde{F_q^{'fake}}) = Distance([\widetilde{f_{s_c}^{'1}}...\widetilde{f_{s_c}^{'T-1}}],[\widetilde{f_q^{'1}}...\widetilde{f_q^{'T-1}}])
\end{equation}
\end{small}

The final distance can be computed as follows:
\begin{small}
\begin{equation} \label{eq13}
d_{all}(s_c,q) = d(\widetilde{F_{s_c}}, \widetilde{F_q^{fake}}) + \alpha d(\widetilde{F_{s_c}^{'}}, \widetilde{F^{'fake}_q})
\end{equation}
\end{small}
where $\alpha$ is a hyper-parameter of motion weight. 

\textbf{Loss.}
We assume $L_{adapt}$ is the cross-entropy loss over the $p^{text-video}(y=i|v)$ and the ground truth, and $L_{all}$ is the cross-entropy loss over the distance $d_{all}$ in \cref{eq13} between the support and query based on the ground truth.
Also, we have the Consistency Loss $L_{con}$ in \cref{eq4} between the fake and real features.
In our work, we have an end-to-end training process, and the final loss can be denoted as:
\begin{small}
\begin{equation}  \label{eq14}
 L=  \lambda_1 L_{adapt} + \lambda_2L_{all} + \lambda_3L_{con}
\end{equation}
\end{small}
where $\lambda_x, x \in \{1,2,3\}$ is the hyper-parameter of weight.  

\subsection{Inference}
In the few-shot action recognition inference stage, we only use the $d_{all}$ for inference. Refer to the \cref{eq13}. The inference can be defined as the \cref{eq15}.
\begin{small}
\begin{equation}  \label{eq15}
    \setlength\belowdisplayskip{-1pt}
    p(y=c|q)=\frac{exp(d_{all}(s_c, q))} {\sum_{\hat{c}=1}^{N} exp(d_{all}(s_{\hat{c}}, q)}
\end{equation}
\end{small}
\section{Experiments}
\subsection{Datasets}
To evaluate our model, we design and conduct the experiments on five datasets which are used for few-shot action recognition, including Kinetics400\cite{carreira2017quo-kinetics}, SSv2-Full\cite{goyal2017something}, SSv2-small\cite{goyal2017something},  UCF101\cite{soomro2012ucf101} and HMDB51\cite{kuehne2011hmdb}. 
For SSv2-small and Kinesics, we use the split method of CMN\cite{zhu2018compound}, CMN-J\cite{zhu2020label}. The split method randomly selects a mini-dataset containing 100 classes, including 64 classes for training, 12 for validation, and 24 for testing, of which each class consists of 100 samples. 
For SSv2-Full, we follow the OTAM\cite{cao2020few-otam} split method. The difference between SSv2-Full and SSv2-small is the former uses all samples of a class. This method contains $77500/1926/2854$ videos for $train/val/test$ respectively.
The HMDB51 contains 31 classes for training, 10 classes for validation, and 10 classes for testing, with $4280/1194/1292$ videos for $train/val/test$.
For the UCF101, there are 70 classes for training, 10 classes for validation, and 21 classes for testing. It contains $9154/1421/2745$ videos for $train/val/test$ respectively. 
The split methods of HMDB51 and UCF101 follow ARN\cite{zhang2020few-ARN}.

\subsection{Implementation Details}
We follow the previous work TSN \cite{wang2016temporal-tsn} for the video frame. 8 frames are sparsely and uniformly sampled from each video. During training, we flip each frame horizontally and randomly crop the center region of 224 × 224 for data augmentation. For the Vision Language Model-CLIP, we use both the version encoder with ResNet50 and VIT-B/16. The optimizer we selected is Adam \cite{2014Adam}, and the learning rate is 0.00001. For training, we thus average gradients and backpropagate once every 16 iterations. For testing, we use only the center crop to augment the video. For inference, there are 10,000 episodes, and the average accuracy of this episode is reported in our experiment. We use the OTAM \cite{cao2020few-otam} as comparison methods for the experiments.

\subsection{Comparison with previous works}
We compare our model with the state-of-the-art methods, and our baseline is the CLIP-FSAR.

\begin{table*}[htbp]
  \centering
  \caption{Comparison on UCF101 and HMDB51. Accuracies for 5-way, 3-shot, and 1-shot settings are shown. $\diamond$ means our implementation
  }
  \vspace{-0.4em}
  \textbf{}
    \label{table:1}
    \scalebox{0.66}{
    \begin{tabular}{l|l|c|l|ccc|ccc}
    \toprule
    \multicolumn{1}{c|}{\multirow{2}[2]{*}{Method}} & \multicolumn{1}{c|}{\multirow{2}[2]{*}{Modal Type}} & \multicolumn{1}{c|}{\multirow{2}[2]{*}{Reference}} & \multicolumn{1}{c|}{\multirow{2}[2]{*}{Backbone}} & \multicolumn{3}{c|}{HMDB51} & \multicolumn{3}{c}{UCF101} \\
          &       &       &       & \multicolumn{1}{c}{1-shot} & \multicolumn{1}{c}{3-shot} & \multicolumn{1}{c|}{5-shot} & \multicolumn{1}{c}{1-shot} & \multicolumn{1}{c}{3-shot} & \multicolumn{1}{c}{5-shot} \\
    \midrule
    ProtoNet \cite{snell2017prototypical_few_shot_learning2}& mono-modal&NeurIPS’2017&ResNet-50& 54.2&- &   68.4& 70.4& - &89.6 \\
    OTAM\cite{cao2020few-otam}  &mono-modal&CVPR’2020&ResNet-50&54.5 & - &66.1&79.9&- &88.9\\
    TRX \cite{perrett2021temporal-trx} &mono-modal&CVPR’2021&ResNet-50& $52.9^{\diamond}$ &- &75.6 &$77.3^{\diamond}$&-&96.1 \\
    STRM \cite{thatipelli2022spatio-strm}&mono-modal &CVPR’2022&ResNet-50& $54.1^{\diamond}$ &- &77.3&$79.2^{\diamond}$& - &96.9  \\
    HyRSM\cite{wang2022-hybrid}&mono-modal &CVPR’2022&ResNet-50&60.3  &71.7 & 76.0 & 83.9 & 93.0 & 94.7 \\
    MTFAN\cite{wu2022motion}&mono-modal&CVPR’2022&ResNet-50& 59.0 &- & 74.6 &84.8 & -& 95.1 \\
    TA2N\cite{li2022ta2n}&mono-modal&AAAI’2022&ResNet-50& 59.7 & - &73.9 &81.9  & -&95.1  \\
    HCL\cite{zheng2022few-hcl}&mono-modal&ECCV’2022&ResNet-50& 59.1 & - &76.3 &82.6  & -&94.5  \\
    TADRNet\cite{2023Task-AwareDual-Representation}&mono-modal&TCSVT'2023&ResNet-50&64.3&74.5&78.2&86.7&94.3&96.4 \\
    MoLo\cite{wang2023molo} &mono-modal&CVPR'2023&ResNet-50& 60.8&72.0&77.4& 86.0&93.5&95.5  \\
    AMeFu-Net(2020)\cite{fu2020depth}&multi-modal&MM’2020&ResNet-50&60.2 &- &75.5& 85.1&-&95.5\\
    SRPN(2021)\cite{wang2021semantic}& multi-modal &MM'2021&ResNet-50 &61.6 & 72.5 &76.2 & 86.5 & 93.8  & 95.8   \\
    TAda-Net(2022)\cite{wang2022task}   & multi-modal& MMSP'2022&ResNet-50&60.8 & 71.8 &76.4  & 85.7 & 93.3 &95.7  \\
    AMFAR(2023)\cite{wanyan2023active}  & multi-modal&CVPR'2023&ResNet-50& 73.9 &- & 87.8& 91.2 &- & 99.0 \\
    CLIP-FSAR(2023)\cite{wang2023CLIP}& multi-modal&IJCV'2023&CLIP-RN50 &  69.4  & 78.3 & 80.7  &92.4 & 95.4 &97.0 \\ 
    CLIP-FSAR(2023)\cite{wang2023CLIP}& multi-modal&IJCV'2023&CLIP-VIT-B &  77.1  & 84.1 & 87.7  &97.0 & 98.5 & 99.1 \\  
    \midrule
\rowcolor{gray!15}    CLIP-CP$\mathrm{M^2}$C      &multi-modal&-&CLIP-RN50& 66.3 & 78.4 $\uparrow$  & 81.2$\uparrow$ & 91.1& 96.2$\uparrow$& 97.4$\uparrow$  \\
\rowcolor{gray!15}    CLIP-CP$\mathrm{M^2}$C      &multi-modal&-&CLIP-VIT-B& 75.9  & 84.4 $\uparrow$ & 88.0$\uparrow$ &  95.0 &  98.1     & 98.6  \\
    \bottomrule
    \end{tabular}%
    }
\end{table*}%

\textbf{Result of UCF101 and HMDB51.}
The standard 5-way 1-shot, 3-shot, and 5-shot settings are scheduled in the experiment.
\cref{table:1} shows the comparison with various state-of-the-art methods. 
\textbf{HMDB51}:
Compared with CLIP-FSAR(RN50), our CLIP-CP$\mathrm{M^2}$C(RN50) brings
$0.5\% (80.7\% \rightarrow 81.2\%)$, 
$0.1\% (78.3\% \rightarrow  78.4\%)$ improvements under the 5-shot and 3-shot settings. 
Compared with CLIP-FSAR(VIT), our CLIP-CP$\mathrm{M^2}$C(VIT) brings
$0.3\% (87.7\% \rightarrow  88.0\%)$, 
$0.3\% (84.1\% \rightarrow 84.4\%)$ improvements under the 5-shot and 3-shot settings. 
\textbf{UCF101}: 
compared with CLIP-FSAR(RN50), our CLIP-CP$\mathrm{M^2}$C(RN50) brings
$0.4\% (97.0\% \rightarrow  97.4\%)$, 
$0.8\% (95.4\% \rightarrow  96.2\%)$ gains under the 5-shot and 3-shot settings.

\textbf{Results of Kinetics and SSv2.}
\cref{table:2} shows the comparison with various methods. 
\textbf{Kinetics}:
compared with CLIP-FSAR(RN50), our CLIP-CP$\mathrm{M^2}$C(RN50) brings 
$0.7\% (90.8\% \rightarrow 91.5\%)$,
$1.1\% (92.0\% \rightarrow 93.1\%)$ 
performance improvements for the 3-shot and 5-shot.
We can see under the 3-shot and 5-shot settings, CLIP-FSAR(RN50) can achieve the highest accuracy among the RN50-based methods.
Except for the 1-shot settings, our CLIP-CP$\mathrm{M^2}$C(VIT) achieves the highest accuracy.
\textbf{SSv2-Full}:
compared with CLIP-FSAR(RN50), our CLIP-CP$\mathrm{M^2}$C(RN50) brings 
$1.1\% (60.7\% \rightarrow 61.8\%)$, 
$1.2\% (62.8\% \rightarrow 64.0\%)$ performance improvements for the 3-shot and 5-shot.
The accuracy of CLIP-CP$\mathrm{M^2}$C(VIT) under 3-shot and 5-shot is higher than CLIP-FSAR(VIT).
\textbf{SSv2-small}: 
The accuracy of CLIP-CP$\mathrm{M^2}$C(RN50) is 
$0.8\% (54.0\% \rightarrow 54.8\%)$ higher than the CLIP-FSAR(RN50) under the 3-shot, 
and $1.3\% (55.8\% \rightarrow 57.1\%)$ higher under 5-shot.
The accuracy is the highest among the RN50-based models.
The accuracy of CLIP-CP$\mathrm{M^2}$C(VIT) under 5-shot is higher than CLIP-FSAR(VIT) and the 3-shot accuracy is a little lower.

Except for the 1-shot setting, our CLIP-CP$\mathrm{M^2}$C is leading for other models and achieves \textbf{state-of-the-art} under the Kinetics, HMDB51, SSv2-small, and SSv2-Full. We can see that the accuracies of 1-shot are slightly lower than CLIP-FSAR. This reason may be due to the fact that under the 1-shot setting, the negative impact of noise caused by the introduction of the fake-labeled query is greater than the positive contribution of using the fake-labeled query to enhance the amount of information. But our 1-shot results are still higher than most of the methods. 

\begin{table*}[htbp]
  \centering
  \caption{Comparison on Kinetics and SSv2 datasets. Accuracy results for 5-way, 3-shot, and 1-shot settings are shown. $\diamond$ means the results of our implementation (the data in parentheses represents the published data).}
  \vspace{-0.4em}
    \label{table:2}
    \scalebox{0.66}{
    \begin{tabular}{l|l|c|l|ccc|ccc|ccc}
    \toprule
    \multicolumn{1}{c|}{\multirow{2}[2]{*}{Method}} & \multicolumn{1}{c|}{\multirow{2}[2]{*}{Modal Type}} & \multicolumn{1}{c|}{\multirow{2}[2]{*}{Reference}} & \multicolumn{1}{c|}{\multirow{2}[2]{*}{Backbone}} & \multicolumn{3}{c|}{Kinetics} & \multicolumn{3}{c|}{SSv2-Full} & \multicolumn{3}{c}{SSv2-small} \\
          &       &       &       & \multicolumn{1}{c}{1-shot} & \multicolumn{1}{c}{3-shot} & \multicolumn{1}{c|}{5-shot} & \multicolumn{1}{c}{1-shot} & \multicolumn{1}{c}{3-shot} & \multicolumn{1}{c|}{5-shot} & \multicolumn{1}{c}{1-shot} & \multicolumn{1}{c}{3-shot} & \multicolumn{1}{c}{5-shot} \\
    \midrule
    ProtoNet \cite{snell2017prototypical_few_shot_learning2}& mono-modal&NeurIPS’2017&ResNet-50& 65.4 &-&77.9&-& &-&33.6 &- & 43.0 \\
    Matching Net\cite{vinyals2016matching—few_shot_learning5}&mono-modal&NeurIPS’2016 &ResNet-50 &53.3& -& 74.6 &-&-&-& 34.4 & -&43.8 \\
    OTAM\cite{cao2020few-otam}          &mono-modal&CVPR’2020&ResNet-50&73.0 &- &85.5 &42.8 &- & 52.3 &$38.9^{\diamond}$ &- & $48.1^{\diamond}$ \\
    TRX \cite{perrett2021temporal-trx}  &mono-modal&CVPR’2021&ResNet-50& $63.4^{\diamond}$(63.6)  &- & $85.1^{\diamond}$(85.9) &42.0 &-& $63.0^{\diamond}$(64.6)  &36.0 &-& $56.3^{\diamond}$(59.4) \\
    STRM \cite{thatipelli2022spatio-strm}&mono-modal &CVPR’2022&ResNet-50& $65.3^{\diamond}$ & -  &$85.9^{\diamond}$(86.7)  &$42.9^{\diamond}$ &- &$64.8^{\diamond}$(68.1) &  &- &   \\
    HyRSM\cite{wang2022-hybrid}         &mono-modal &CVPR’2022&ResNet-50& 73.7  &83.5 & 86.1 &54.3  & 65.1  &  69.0 & 40.6 & 52.3  &56.1  \\
    MTFAN\cite{wu2022motion}            &mono-modal&CVPR’2022&ResNet-50& 74.6  &- & 87.4 &45.7 &  -  &  60.4 & - & -  &-  \\
    TA2N\cite{li2022ta2n}               &mono-modal&AAAI’2022&ResNet-50&72.8& -& 85.8 & 47.6 & -&61.0  & - & - & -   \\
    HCL\cite{zheng2022few-hcl}               &mono-modal&AAAI’2022&ResNet-50&73.7& -& 85.8 & 47.3 & -&64.9  &38.7& - & 55.4   \\
    TADRNet\cite{2023Task-AwareDual-Representation}&mono-modal&TCSVT'2023&ResNet-50& 75.6 & 84.8 & 87.4  &43.0 &-  &61.1&- & - &- \\
    MoLo\cite{wang2023molo}             &mono-modal &CVPR'2023&ResNet-50&74.0 &83.7& 85.6 &56.6 &67.0& 70.6 & 42.7 &52.9& 56.4 \\
    AMeFu-Net(2020)\cite{fu2020depth}   &multi-modal&MM’2020&ResNet-50& 74.1  &- &86.8 &- &-&- & - &-& - \\
    CMN++(2020)\cite{zhu2020label}      &multi-modal &TRAMI'2020&ResNet-50&60.5 &75.6 &78.9 & 36.2& 44.6 & 44.8 & - &-  &-  \\
    SRPN(2021)\cite{wang2021semantic}   & multi-modal &MM'2021&ResNet-50&75.2 & 84.7 &87.1  & -   & -   & -   & -  &  -  & - \\
    AMFAR(2023)\cite{wanyan2023active}  & multi-modal&CVPR'2023&ResNet-50&80.1 & -& 92.6 &61.7& -& 79.5 & -  & -  &-   \\
    CLIP-FSAR(2023)\cite{wang2023CLIP}  & multi-modal&IJCV'2023&CLIP-RN50 &90.1&90.8&92.0& 58.7&60.7&62.8 & 52.1 &54.0&55.8\\ 
    CLIP-FSAR(2023)\cite{wang2023CLIP}  & multi-modal&IJCV'2023&CLIP-VIT-B &94.8&95.0&95.4& 62.1&68.3&72.1 & 54.6 &59.4&61.8\\ 
    \midrule
    \rowcolor{gray!15}    CLIP-CP$\mathrm{M^2}$C        &multi-modal&-&CLIP-RN50& 88.6 & 91.5 $\uparrow$ & 93.1$\uparrow$ & 58.0 & 61.8 $\uparrow$ & 64.0 $\uparrow$&   51.5 & 54.8 $\uparrow$ & 57.1 $\uparrow$  \\
    \rowcolor{gray!15}    CLIP-CP$\mathrm{M^2}$C        &multi-modal&-&CLIP-VIT-B&91.0 & 95.1 $\uparrow$ & 95.5 $\uparrow$ & 60.1 &  68.9 $\uparrow$ & 72.8 $\uparrow$ & 52.3 &  58.9 &  62.6 $\uparrow$ \\
    \bottomrule
    \end{tabular}%
    }
\end{table*}%

\subsection{Ablation Study}
\textbf{Analysis of Model Components.}
To validate the contributions of each module in our work, we give the experiment under 5-way, 5-shot settings on Kinetics, SSv2-Small, and HMDB51 datasets. The CLIP-RN50 is chosen as the backbone for the visual encoder. As shown in \cref{table:3}, it can be seen that by using CPM with Consistency Loss and also using the normal frame features and the motion features from the MC module, we can get the highest accuracy. For the Kinetics, it brings the accuracy of $93.1\%$. Also, Under the SSv2-Small, it gets $57.1\%$ with all the settings.  
Through the \cref{table:3}, 
we can conclude that 
(1) Consistency Loss can improve the recognition accuracy.
(2) The accuracy is higher using both normal frame features and motion features simultaneously than only using a single type.
(3) Motion feature features can supplement normal features to a certain extent.
(4) The recognition using motion features alone is workable, and the accuracies are acceptable.

\begin{table*}[htbp]
  \centering
  \caption{Ablation for the Analysis of Model Components. It is for CPM(Consistency Loss) and the Feature type we choose.}
  \vspace{-0.4em}
  \label{table:3}
   \scalebox{0.66}{
    \begin{tabular}{c|c|cc|c|c|c}
    \toprule
    \multicolumn{1}{c|}{\multirow{2}[2]{*}{serial No}} & \multicolumn{1}{c|}{CPM} & \multicolumn{2}{c|}{Feature Type} & Kinetics & SSv2-Small & \multicolumn{1}{l}{HMDB51} \\
          & \multicolumn{1}{c|}{(Consistency-loss)} & Normal & Motion & 5-shot & 5-shot & \multicolumn{1}{l}{5-shot} \\
    \midrule
      (1)    & \ding{51}  & \ding{51}& \ding{51}  &93.1  & 57.1 & 81.2 \\
      (2)    & \ding{51}  & \ding{51}& \ding{55}  &92.8  & 55.8 & 81.0 \\
      (3)    & \ding{51}  & \ding{55}& \ding{51}  &91.6  & 56.4 & 80.5 \\
      (4)    & \ding{55}  & \ding{51}& \ding{51}  &92.6  & 56.4 & 80.8 \\
      (5)    & \ding{55}  & \ding{51}& \ding{55}  &92.3  & 55.4 & 80.1 \\
      (6)    & \ding{55}  & \ding{55}& \ding{51}  &90.9  & 55.9 & 79.9 \\
    \bottomrule
    \end{tabular}%
    }
\end{table*}%

\textbf{Analysis of Domain Adaption.}
To check the affection of the Domain Adaption, we use Kinetics and SSv2-Small under 1-shot and 5-shot settings with Consistency Loss for normal and motion features. In \cref{table:4}, the accuracy with Domain Adaption is higher than without it. It indicates that Domain Adaption, similar to many previous CLIP-related works, does contribute to our model.
\begin{table}[htbp]
  \centering
  \caption{Ablation for Domain Adaption in our model.}
  \vspace{-0.4em}
  \label{table:4}
  \scalebox{0.66}{
    \begin{tabular}{c|cc|cc}
    \toprule
    \multirow{2}[2]{*}{Domain-Adaption} & \multicolumn{2}{c|}{Kinetics} & \multicolumn{2}{c}{SSv2-Small} \\
          & 1-shot & 5-shot & 1-shot  & 5-shot \\
    \midrule
      \ding{51} &88.6&93.1& 51.5& 57.1 \\
      \ding{55} &87.3&92.4& 51.0  & 56.0 \\
    \bottomrule
    \end{tabular}%
    }
\end{table}%

\begin{figure}
\setlength{\belowcaptionskip}{-0.5cm} %调整图片标题与下文距离
\centering
\includegraphics[width=7cm]{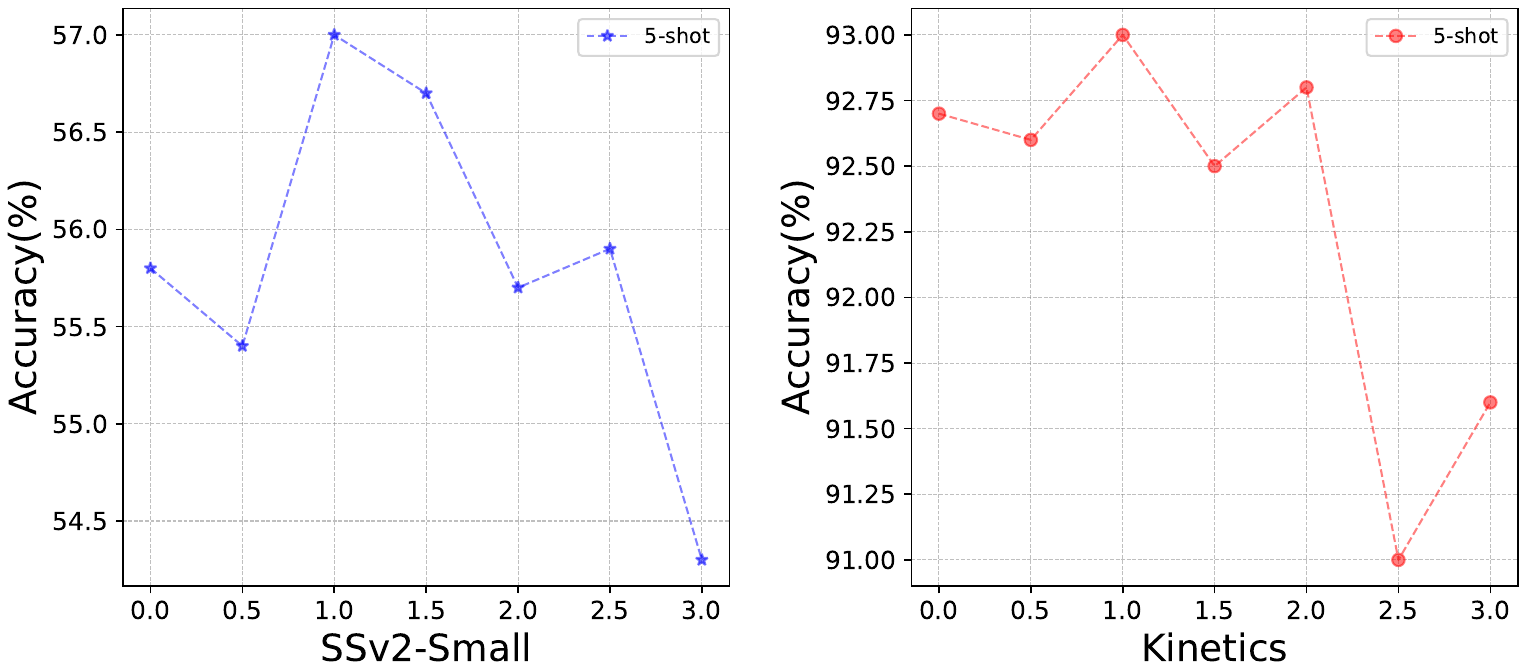}
\caption{\label{fig:parameter}The sensitivity of the hyper-parameter $\alpha$.
} 
\label{fig:3}
\end{figure}

\textbf{Hyper-parameter sensitivity of Motion analysis.}
In our work, a hyper-parameter balances the contribution of motion level and normal frame level distance. In \cref{fig:3}, we can observe that the parameter has an impact on the performance. According to \cref{fig:3}, in the experiments, we keep the hyper-parameter to be $1$.

\textbf{Fusion methods Experiment}
To prove that our Feature Enhancement for Consistency Prototype Module based on Transformer \eqref{subsubsec: feature enhancement} is more
efficient than other modal fusion methods, just as a weighted sum method for visual and text(prompt) embedding or a simple concatenation method. We select the CLIP(RN50) as the backbone and experiment to compare different methods on SSv2-small and Kinetics. See in the \cref{table:5}, we can see that our Feature Enhancement is more efficient than the other methods.

\begin{table}[htbp]
  \centering
  \caption{Comparision between different feature fusion methods}
  \scalebox{0.8}{
    \begin{tabular}{l|c|c|c}
    \toprule
    Method & \multicolumn{1}{l|}{SSv2-small} & \multicolumn{1}{l|}{SSv2-small} & \multicolumn{1}{l}{Kinetics} \\
    \midrule
    weighted sum & 62.9 & 55.1 & 91.3 \\
    concatenation & 63.4& 56.7& 92.8\\
    feature enhancement& 64.0 & 57.1 & 93.1\\
    \bottomrule
    \end{tabular}%
  }
  \label{table:5}%
\end{table}%

\section{Supplementary Visualization}
\label{sec:supp-visulization}

\subsection{Accuracy comparison of different classes}
\begin{figure}[htbp]
    \centering
    \subfloat[Accuracy comparison for Kinetics]{
    \label{fig:4.a}
    \includegraphics[width=8cm]{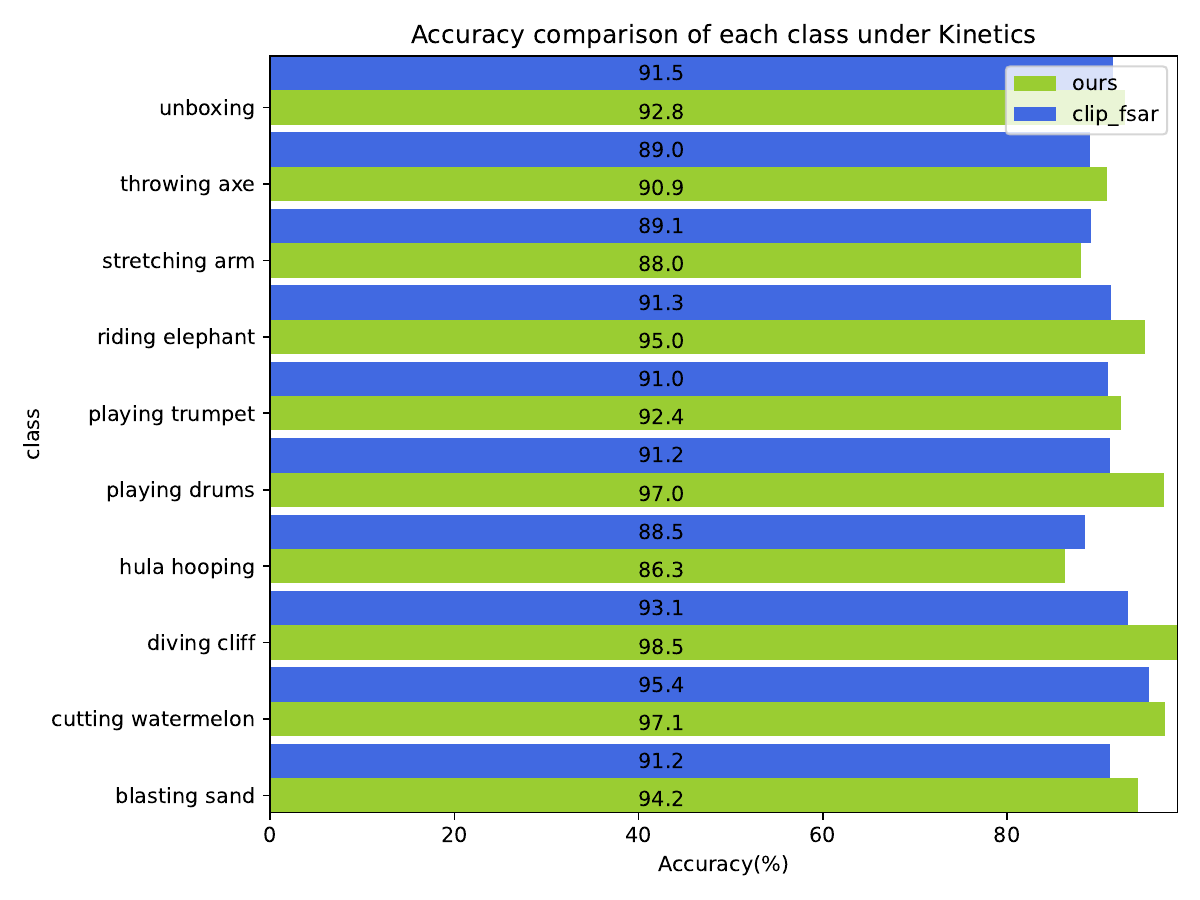}
    }
    \\
    \subfloat[Accuracy comparison for SSv2-small]{
    \label{fig:4.b}
    \includegraphics[width=8cm]{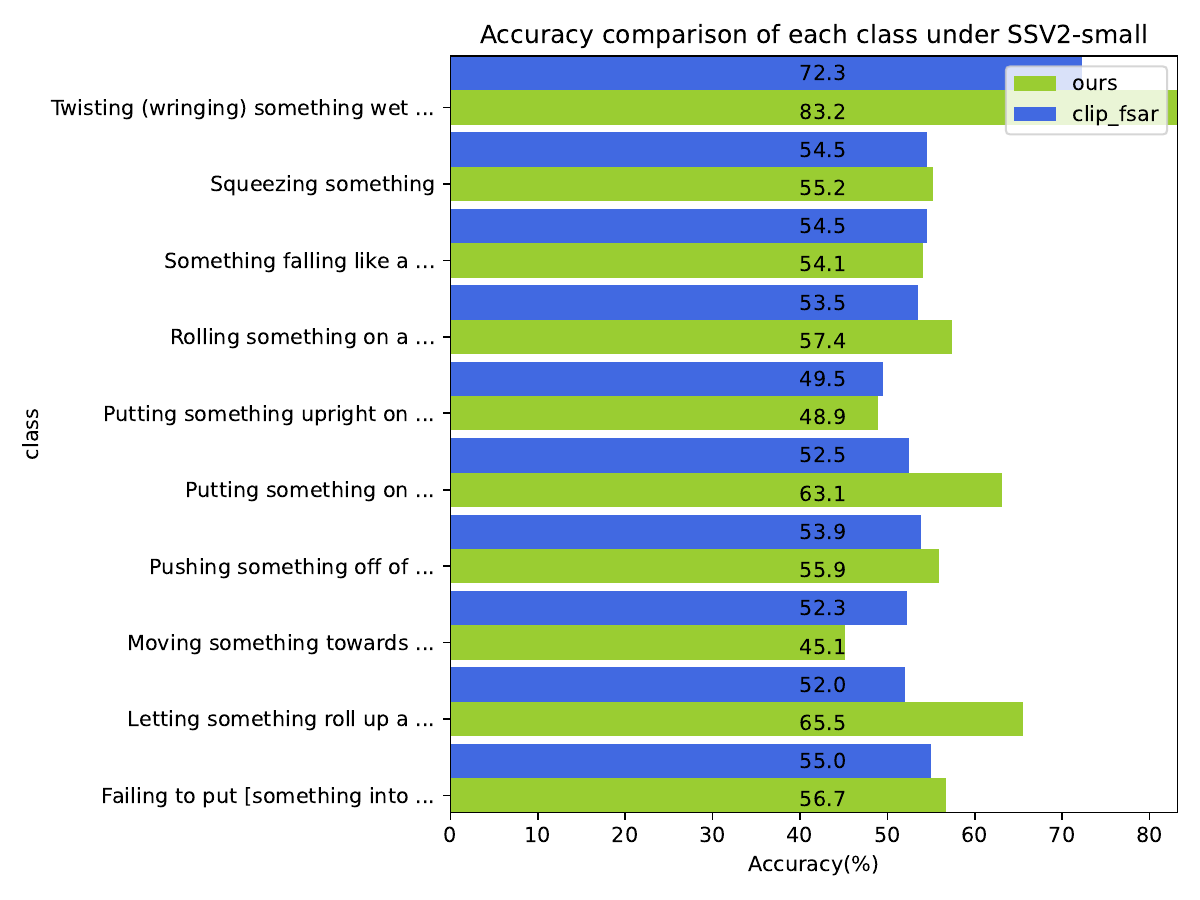}
    }
    \caption{The illustration shows the comparison of CLIP-FSAR and CLIP-CP$\mathrm{M^2}$C for the accuracy of Kinetics and SSv2-small.
    }
\label{fig:4}
\end{figure}

In our comparative study with CLIP-FSAR \cite{wang2023CLIP} on Kinetics, ten classes were randomly selected under the 5-shot setting. The results are illustrated in \cref{fig:4.a} and \cref{fig:4.b}, reveal notable accuracy improvements for our CLIP-CP$\mathrm{M^2}$C model across various classes. Notably, in Kinetics, the accuracy for the "playing drums" class demonstrated the most improvement, rising from $91.2\%$ with CLIP-FSAR to an impressive $97.0\%$ with our model. Similarly, in SSv2-small, for the class "letting something roll up a slanted surface, so it rolls back down," the accuracy increased significantly from $52.0\%$ with CLIP-FSAR to $65.5\%$ with our model. These results demonstrate the effectiveness of our model in enhancing accuracy for specific action classes.

\subsection{Distribution comparison}
\begin{figure*}[htbp]
    \centering
    \subfloat[Ours-Normal]{
    \label{fig:5.a}
    \includegraphics[width=4cm]{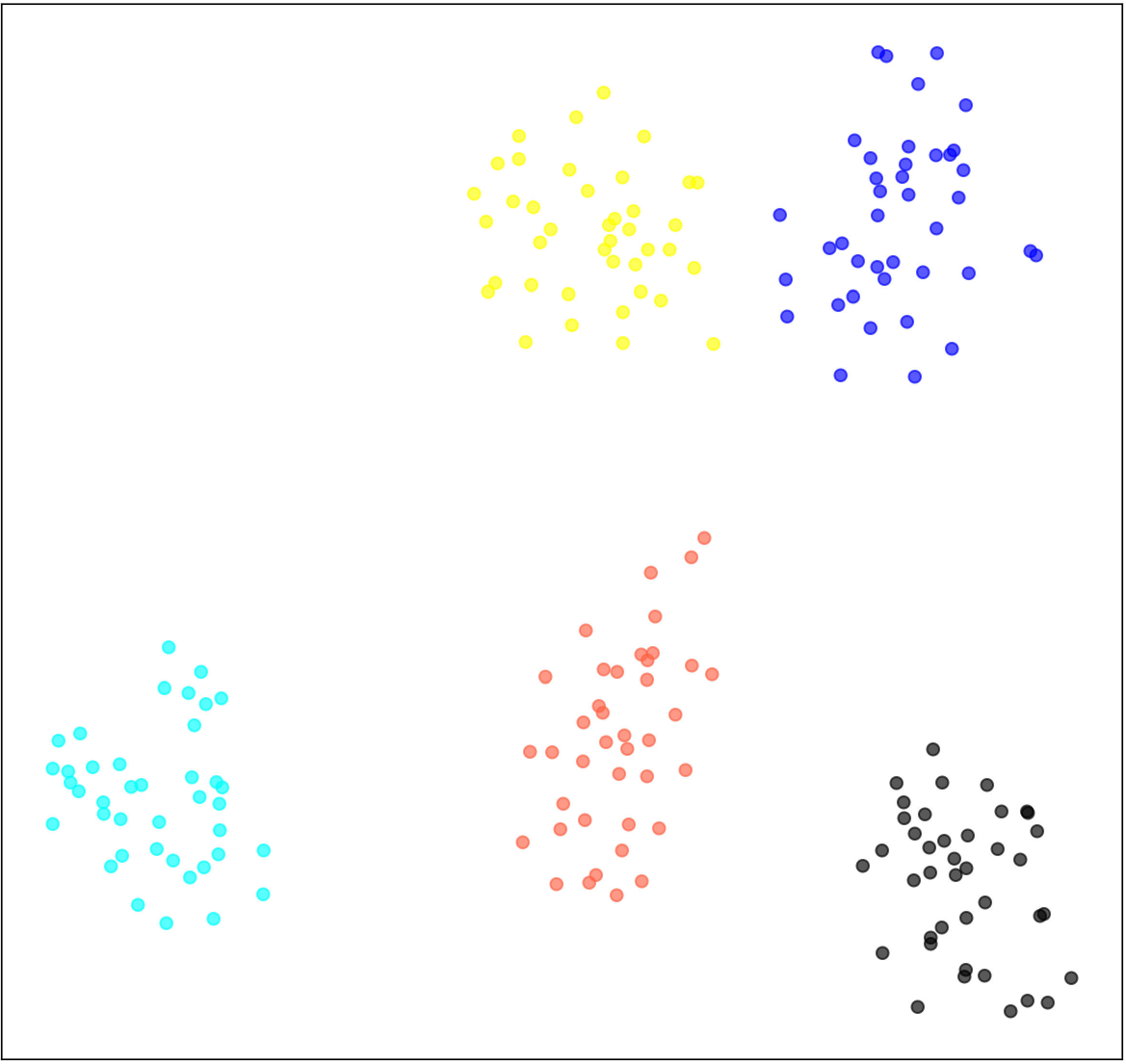}
    }
    \subfloat[Ours-Motion]{
    \label{fig:5.b}
    \includegraphics[width=4cm]{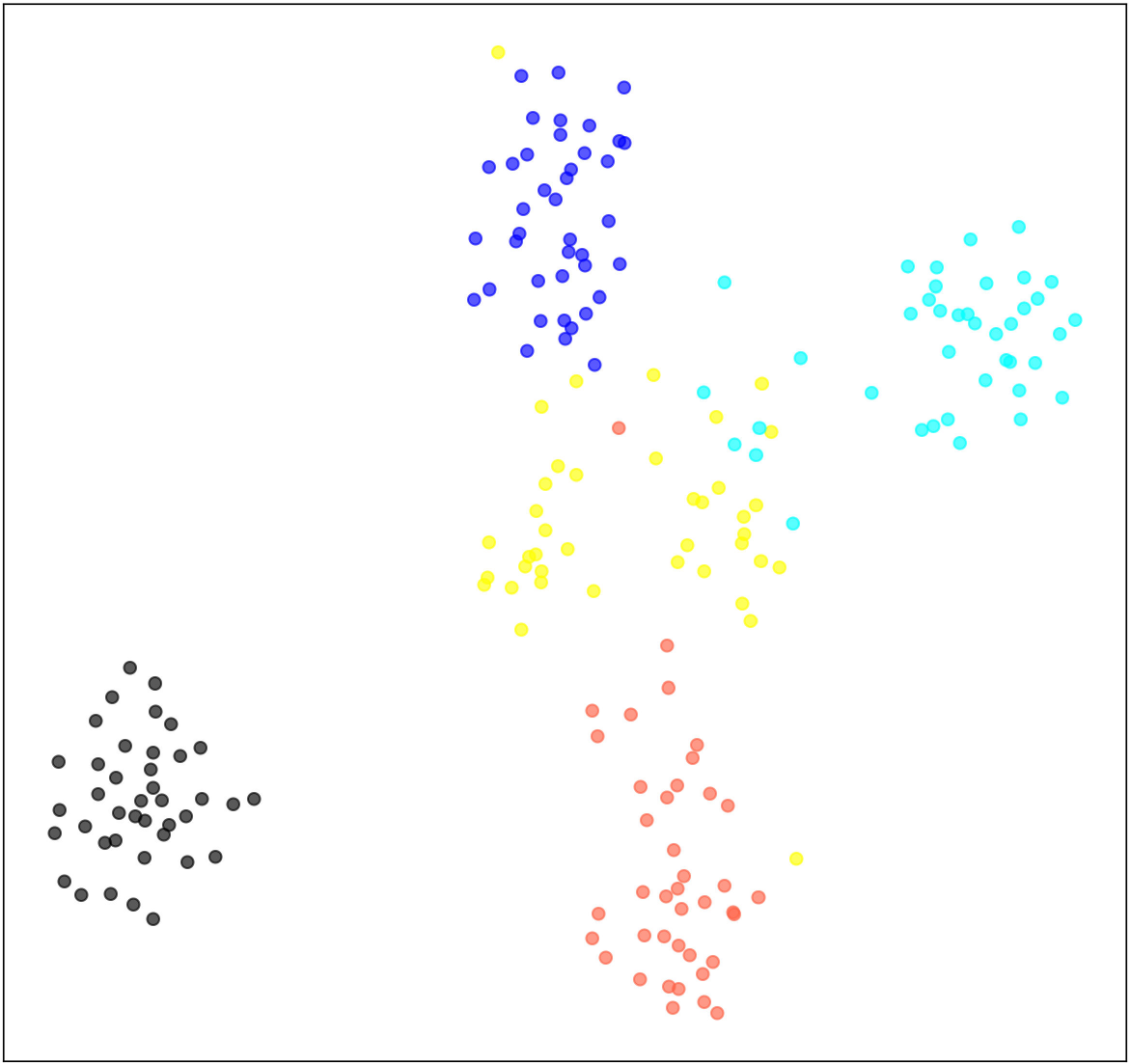}
    }
    \subfloat[CLIP-FSAR]{
    \label{fig:5.c}
    \includegraphics[width=4cm]{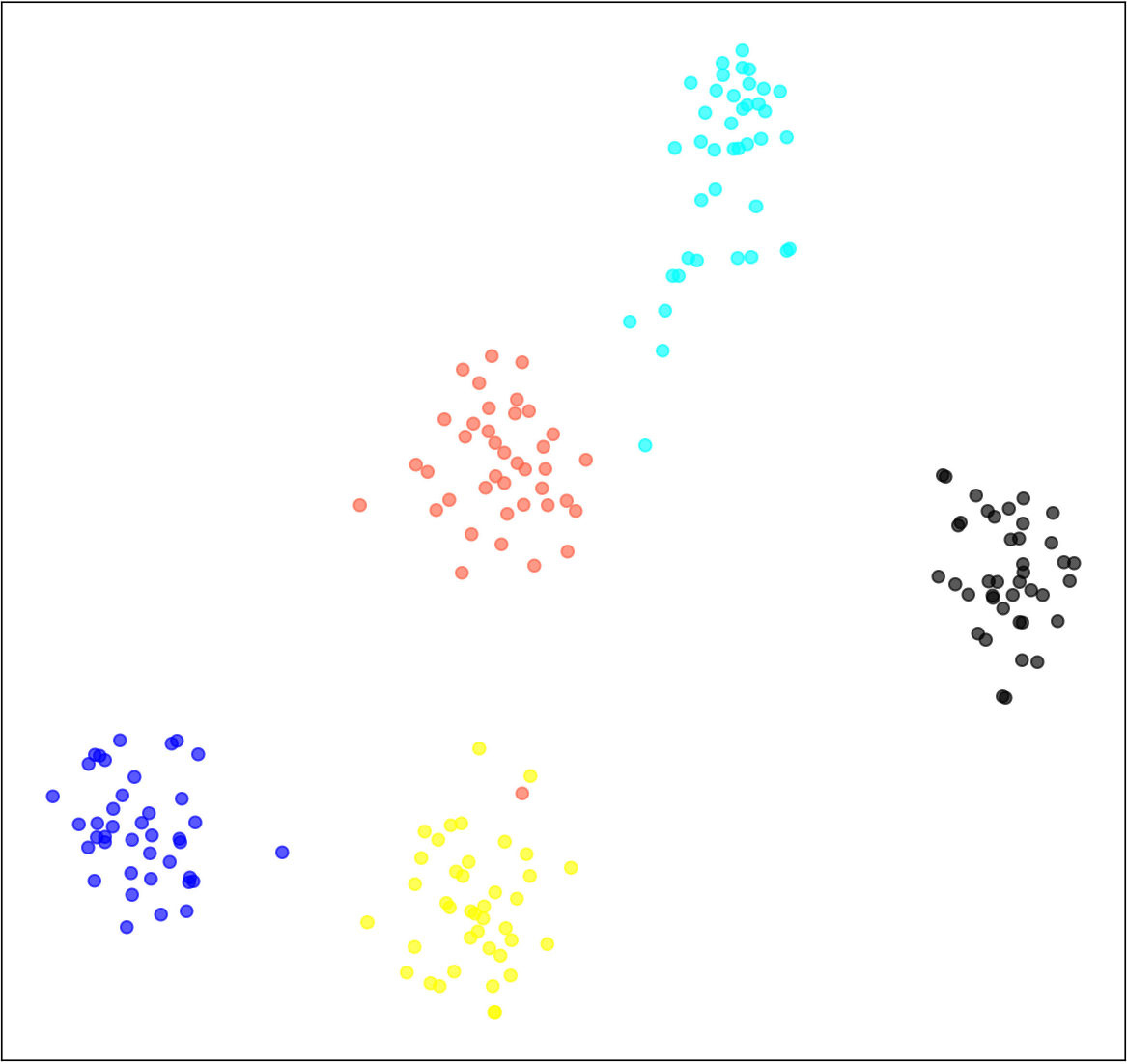}
    }
    \caption{Feature distribution under Kinetics.
    }
\label{fig:5}
\end{figure*}

\begin{figure*}[htbp]
    \centering
    \subfloat[Ours-Normal]{
    \label{fig:6.a}
    \includegraphics[width=4cm]{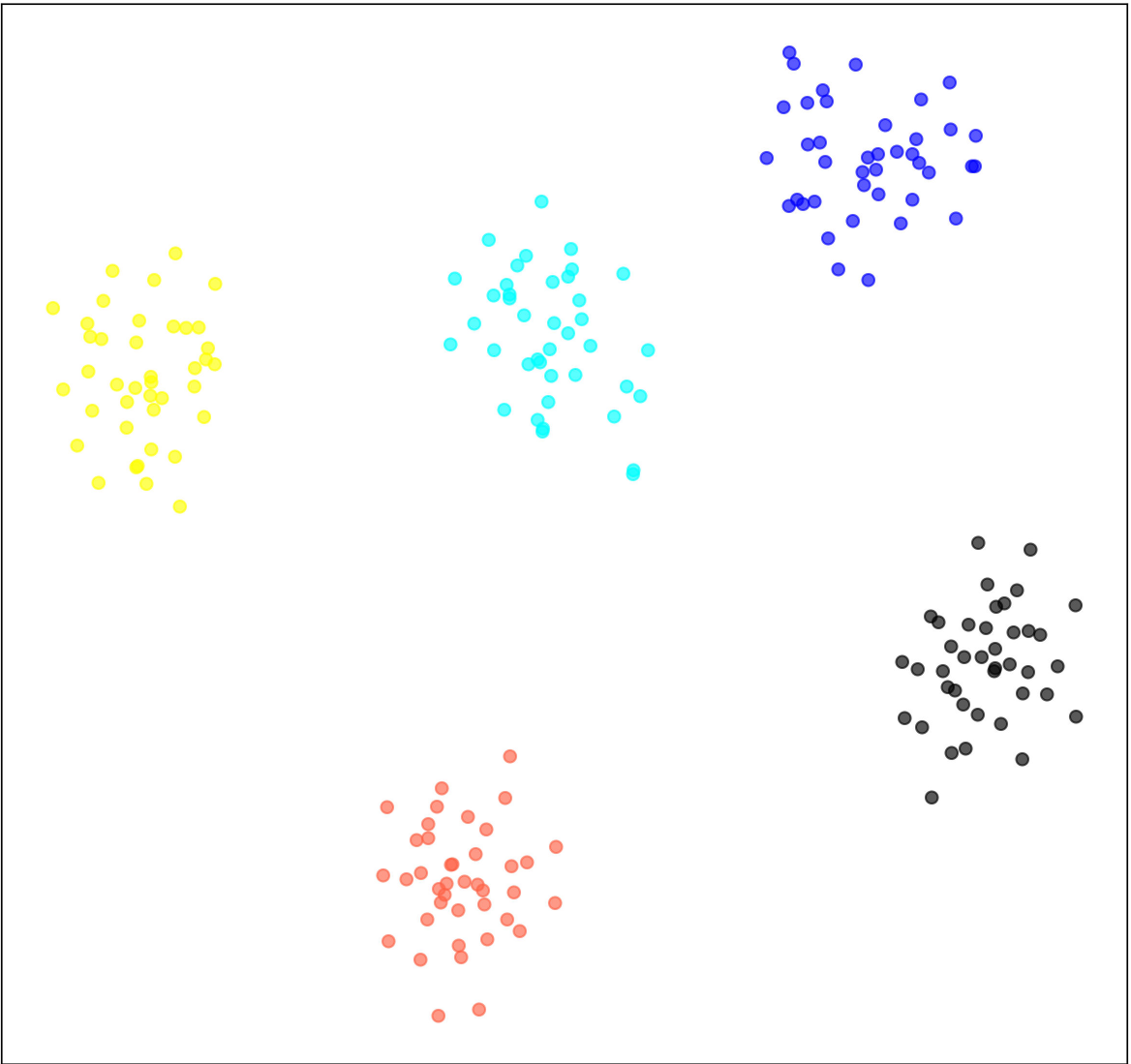}
    }
    \subfloat[Ours-Motion]{
    \label{fig:6.b}
    \includegraphics[width=4cm]{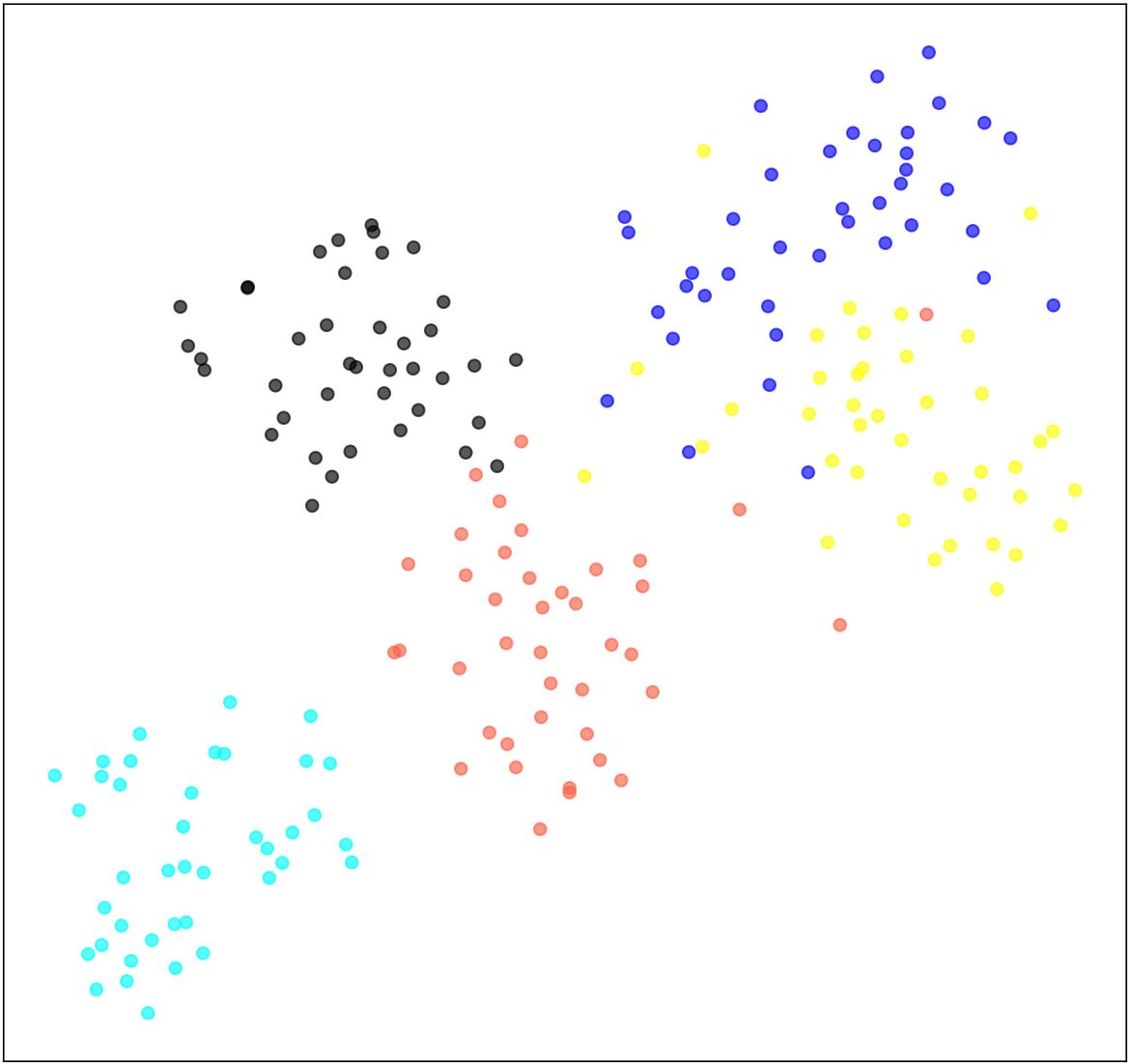}
    }
    \subfloat[CLIP-FSAR]{
    \label{fig:6.c}
    \includegraphics[width=4cm]{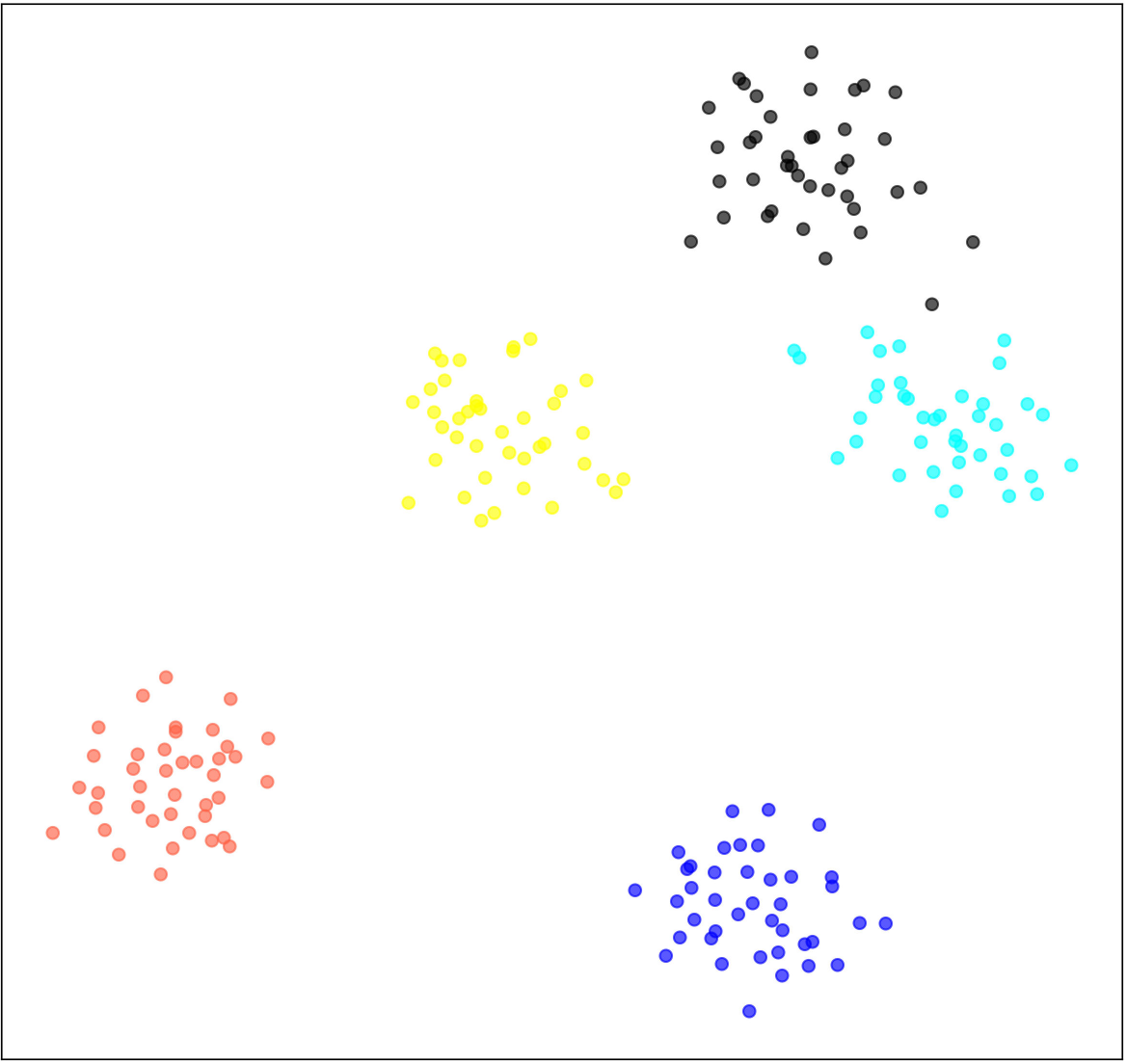}
    }
    \caption{Feature distribution under SSv2-small.
    }
\label{fig:6}
\end{figure*}

To validate the superior data distribution of CLIP-CP$\mathrm{M^2}$C, we conduct t-SNE \cite{wang2016temporal-tsn} visualizations on Kinetics and SSv2-small under the 5-way 5-shot setting, as depicted in \cref{fig:5} and \cref{fig:6}. 
We plot the distribution of our model for both the normal frame and motion features.
The visualizations confirm that our model exhibits enhanced discriminative features:
(1) Intra-class distribution of features is more compact. 
(2) Inter-class distribution of features is more discriminative. 
(3) The class distribution of features for the motion feature is also discriminative.


\subsection{Analysis of attention visualization}
\begin{figure}[htbp]
    \centering
    \subfloat[Origin Frames]{
    \label{fig:7.a}
    \includegraphics[width=8cm]{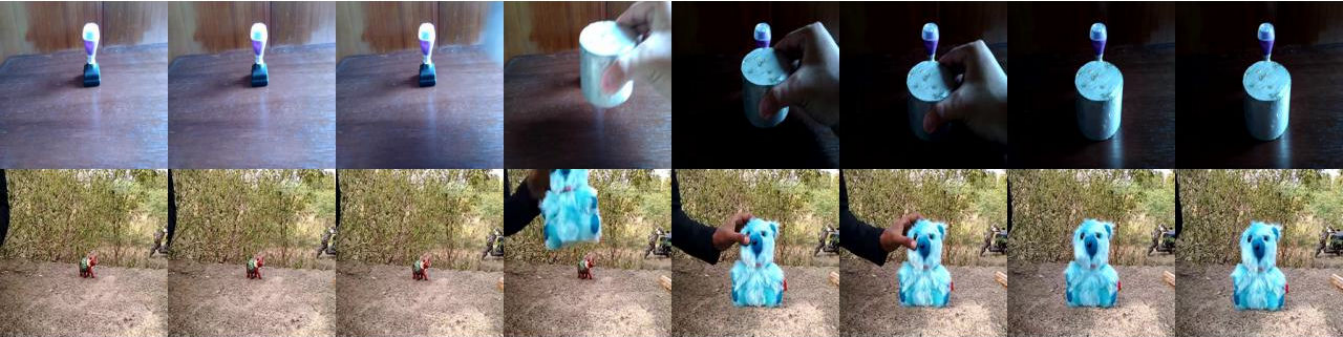}
    }
    \\
    \subfloat[CLIP-CP$\mathrm{M^2}$C(RN50) without Consistency Loss]{
    \label{fig:7.b}
    \includegraphics[width=8cm]{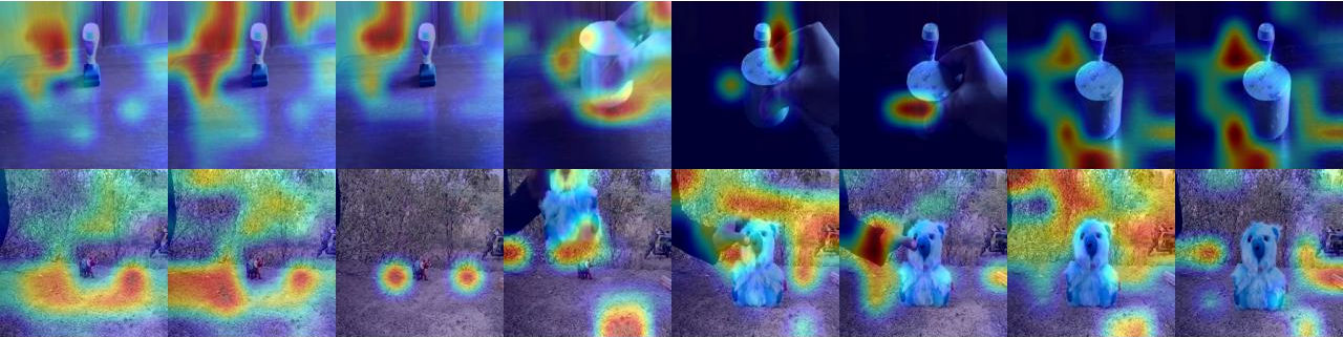}
    }
    \\
    \subfloat[CLIP-CP$\mathrm{M^2}$C(RN50) with Consistency Loss]{
    \label{fig:7.c}
    \includegraphics[width=8cm]{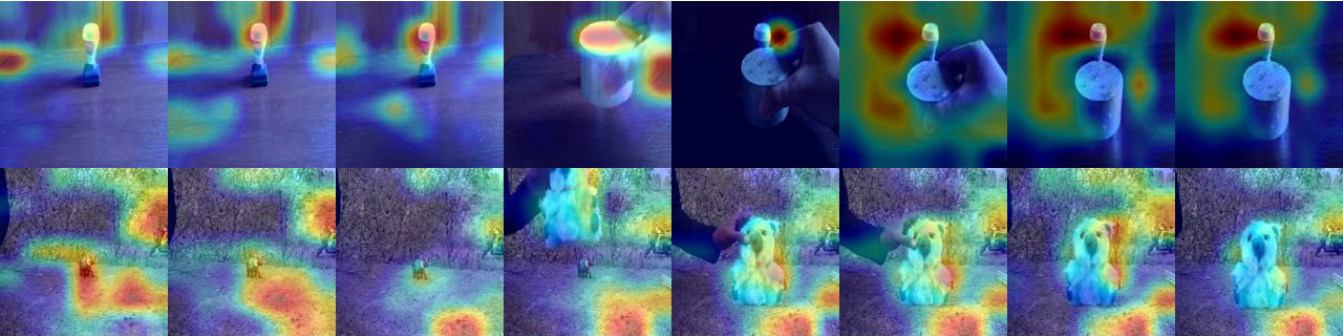}
    }
    \label{fig:attention_map_ssv}
    \caption{Attention visualization of our CLIP-CP$\mathrm{M^2}$C for SSv2-small based on CLIP(RN50). 
    }
\label{fig:7}
\end{figure}

\begin{figure}[htbp]
    \centering
    \subfloat[Origin Frames]{
    \label{fig:8.a}
    \includegraphics[width=8cm]{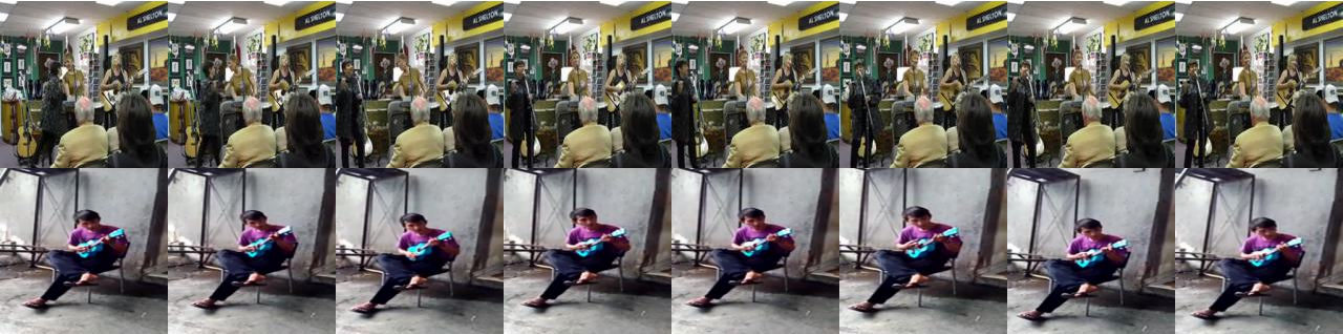}
    }
    \\
    \subfloat[CLIP-CP$\mathrm{M^2}$C (RN50) without Consistency Loss]{
    \label{fig:8.b}
    \includegraphics[width=8cm]{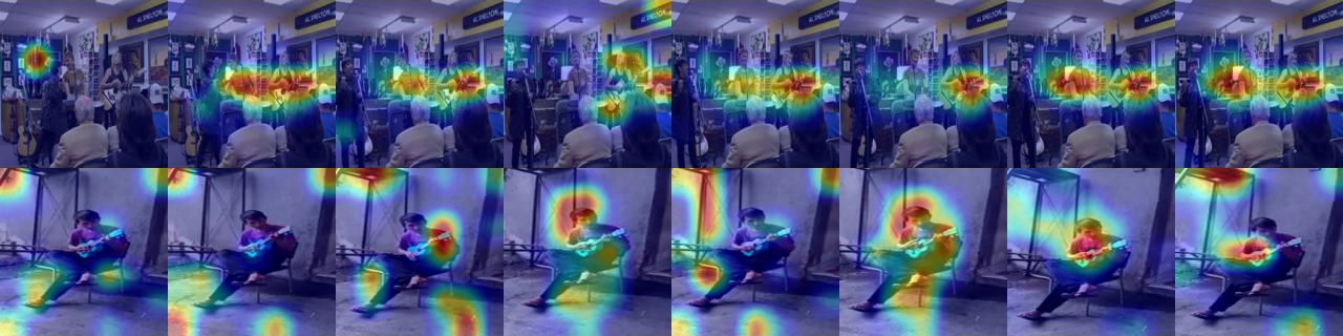}
    }
    \\
    \subfloat[CLIP-CP$\mathrm{M^2}$C (RN50) with Consistency Loss]{
    \label{fig:8.c}
    \includegraphics[width=8cm]{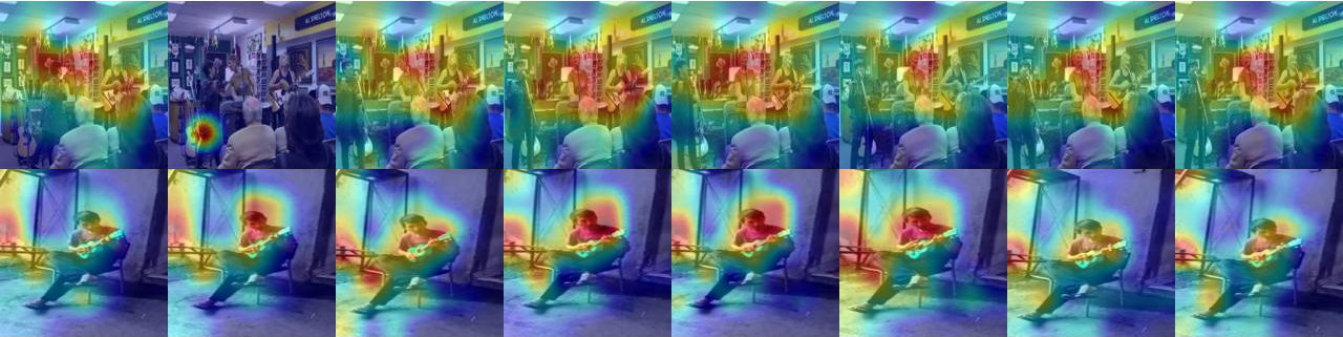}
    }
    \label{fig:attention_map_kinetics}
    \caption{Attention visualization of our CLIP-CP$\mathrm{M^2}$C for Kinetics based on CLIP(RN50). 
    }
\label{fig:8}
\end{figure}

\begin{figure}[htbp]
    \centering
    \subfloat[Origin Frames]{
    \label{fig:9.a}
        \includegraphics[width=8cm]
        {ssv2_putting_origin.pdf}
    }
    \\
    \subfloat[CLIP-CP$\mathrm{M^2}$C (VIT-B) without Consistency Loss]{
    \label{fig:9.b}
        \includegraphics[width=8cm]{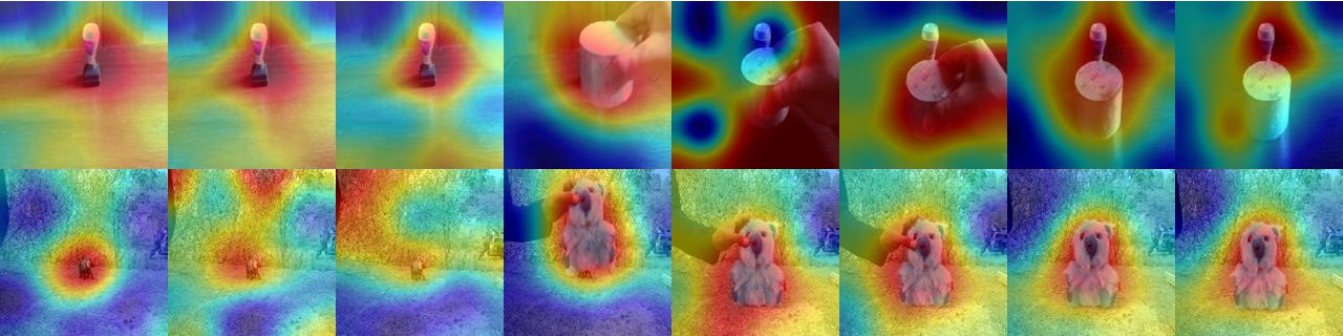}
    }
    \\
    \subfloat[CLIP-CP$\mathrm{M^2}$C (VIT-B) with Consistency Loss]{
    \label{fig:9.c}
         \includegraphics[width=8cm]{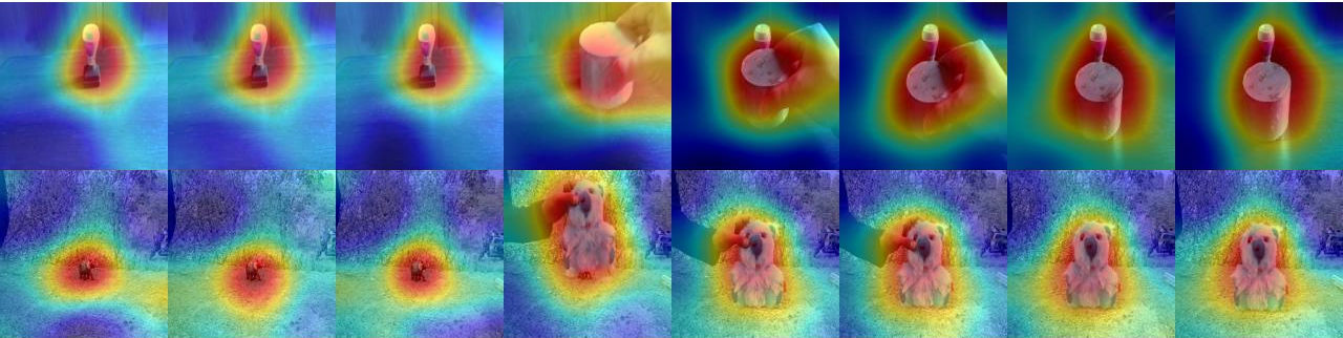}
    }
    \label{fig:attention_map_ssv_vit}
    \caption{Attention visualization of our CLIP-CP$\mathrm{M^2}$C for SSv2-small based on CLIP(VIT-B). 
    }
\label{fig:9}
\end{figure}

\begin{figure}[htbp]
    \centering
    \subfloat[Origin Frames]{
    \label{fig:10.a}
    \includegraphics[width=8cm]{kinetics_play_ukelele_origin.pdf}
    }
    \\
    \subfloat[CLIP-CP$\mathrm{M^2}$C (VIT-B) without Consistency Loss]{
    \label{fig:10.b}
    \includegraphics[width=8cm]{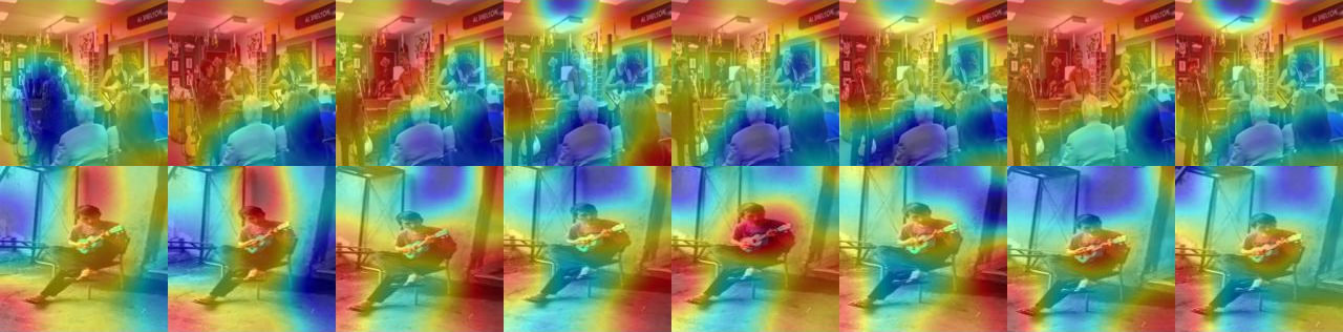}
    }
    \\
    \subfloat[CLIP-CP$\mathrm{M^2}$C (VIT-B) with Consistency Loss]{
    \label{fig:10.c}
    \includegraphics[width=8cm]{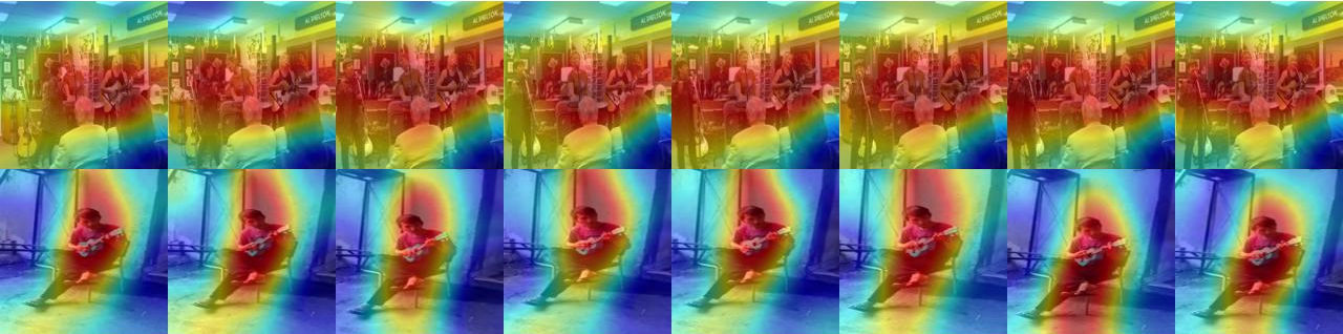}
    }
    \label{fig:attention_map_kinetics_vit}
    \caption{Attention visualization of our CLIP-CP$\mathrm{M^2}$C for Kinetics based on CLIP(VIT-B). 
    }
\label{fig:10}
\end{figure}

For further insight into our model's capabilities, attention visualizations are performed using CLIP(RN50) and CLIP(VIT-B) as the backbone. 
From \cref{fig:7} to \cref{fig:10}, 
based on the original RGB images(upper pair), the attention visualizations of the CLIP-CP$\mathrm{M^2}$C without the Consistency Loss (middle pair) are compared to the attention visualizations with the Consistency Loss(bottom pair) in each figure. For each pair, the first is the support, and the second one is the query.

\textbf{Attention under the CLIP(RN50)}.
For the action category ``Putting something in front of something" in SSv2-small,
the visualizations in \cref{fig:7} reveal that CLIP-CP$\mathrm{M^2}$C effectively focuses on relevant objects and related backgrounds, showcasing its ability to enhance semantic and spatiotemporal representation. 
\cref{fig:8} shows the attention visualizations of CLIP-CP$\mathrm{M^2}$C on Kinetics and the action category is ``playing ukulele". The explanation is similar to what is for \cref{fig:7}

%%%%%%%%%%%%%%%%%%%%%%%%%%%%%%%%%%%%%%%%%%%%%%%%%%%%%%%%%%%%%%%%%%%%%%%%%%%%%%%%%%%%%%%%%%%%%%%%%%%%%%%%%%%%%%%%%%%%%%%%%%%%%%%%%%%%%%%%%%%%%%%%%
\textbf{Attention under the CLIP(VIT-B)}.
\cref{fig:9} shows the attention visualization of our CLIP-CP$\mathrm{M^2}$C on SSv2-small under the 5-way 5-shot setting.
The action class is ``Putting something in front of something", and the backbone is CLIP(VIT-B).
\cref{fig:10} shows the attention visualization of our CLIP-CP$\mathrm{M^2}$C on Kinetics under the 5-way 5-shot setting.
The action class is ``playing ukulele", and the backbone is CLIP(VIT-B).

The model without the Consistency Loss in some frames pays too much attention to unrelated backgrounds.
It can be seen with the Consistency Loss the attention is more accurate for the main objects in most of the frames.

\section{Conclusion}
Based on CLIP, we propose a multi-modal few-shot action model named CLIP-CP$\mathrm{M^2}$C. 
In the meta-training, we concatenate a random vector with frames and also concatenate the real prompt embedding with frames. They are sent to the CPM. The L2 distance is used for the output as the Consistency Loss.
In the meta-testing, we use a random vector for query enhancement and the real prompt embedding of support for the prototype.
With the CNN and difference operation, we obtain the motion features of the video that act as a supplement to the frames. 
During meta-training, we use the text-image cosine similarity to give the model a Domain Adaption.
{
    \small
    \bibliographystyle{ieeenat_fullname}
    \bibliography{main}

\begin{thebibliography}{60}
\providecommand{\natexlab}[1]{#1}
\providecommand{\url}[1]{\texttt{#1}}
\expandafter\ifx\csname urlstyle\endcsname\relax
  \providecommand{\doi}[1]{doi: #1}\else
  \providecommand{\doi}{doi: \begingroup \urlstyle{rm}\Url}\fi

\bibitem[Bishay et~al.(2019)Bishay, Zoumpourlis, and Patras]{bishay2019tarn}
Mina Bishay, Georgios Zoumpourlis, and Ioannis Patras.
\newblock Tarn: Temporal attentive relation network for few-shot and zero-shot action recognition.
\newblock \emph{arXiv preprint arXiv:1907.09021}, 2019.

\bibitem[Cao et~al.(2020)Cao, Ji, Cao, Chang, and Niebles]{cao2020few-otam}
Kaidi Cao, Jingwei Ji, Zhangjie Cao, Chien-Yi Chang, and Juan~Carlos Niebles.
\newblock Few-shot video classification via temporal alignment.
\newblock In \emph{Proceedings of the IEEE/CVF Conference on Computer Vision and Pattern Recognition}, pages 10618--10627, 2020.

\bibitem[Carreira and Zisserman(2017)]{carreira2017quo-kinetics}
Joao Carreira and Andrew Zisserman.
\newblock Quo vadis, action recognition? a new model and the kinetics dataset.
\newblock In \emph{proceedings of the IEEE Conference on Computer Vision and Pattern Recognition}, pages 6299--6308, 2017.

\bibitem[Chen et~al.(2018)Chen, Fu, Zhang, Jiang, Xue, and Sigal]{chen1804semantic-few-augmentation3}
Z Chen, Y Fu, Y Zhang, YG Jiang, X Xue, and L Sigal.
\newblock Semantic feature augmentation in few-shot learning. arxiv 2018.
\newblock \emph{arXiv preprint arXiv:1804.05298}, 2018.

\bibitem[Dang et~al.(2023)Dang, Luo, Jia, Yan, Chang, and Zheng]{10035001-generation-few-shot-learning}
Zhuohang Dang, Minnan Luo, Chengyou Jia, Caixia Yan, Xiaojun Chang, and Qinghua Zheng.
\newblock Counterfactual generation framework for few-shot learning.
\newblock \emph{IEEE Transactions on Circuits and Systems for Video Technology}, pages 1--1, 2023.

\bibitem[Dosovitskiy et~al.(2020)Dosovitskiy, Beyer, Kolesnikov, Weissenborn, Zhai, Unterthiner, Dehghani, Minderer, Heigold, Gelly, et~al.]{vision-transformer}
Alexey Dosovitskiy, Lucas Beyer, Alexander Kolesnikov, Dirk Weissenborn, Xiaohua Zhai, Thomas Unterthiner, Mostafa Dehghani, Matthias Minderer, Georg Heigold, Sylvain Gelly, et~al.
\newblock An image is worth 16x16 words: Transformers for image recognition at scale.
\newblock \emph{arXiv preprint arXiv:2010.11929}, 2020.

\bibitem[Finn et~al.(2017)Finn, Abbeel, and Levine]{finn2017model-MAML-few-shot-model1}
Chelsea Finn, Pieter Abbeel, and Sergey Levine.
\newblock Model-agnostic meta-learning for fast adaptation of deep networks.
\newblock In \emph{International conference on machine learning}, pages 1126--1135. PMLR, 2017.

\bibitem[Fu et~al.(2020)Fu, Zhang, Wang, Fu, and Jiang]{fu2020depth}
Yuqian Fu, Li Zhang, Junke Wang, Yanwei Fu, and Yu-Gang Jiang.
\newblock Depth guided adaptive meta-fusion network for few-shot video recognition.
\newblock In \emph{Proceedings of the 28th ACM International Conference on Multimedia}, pages 1142--1151, 2020.

\bibitem[Goyal et~al.(2017)Goyal, Ebrahimi~Kahou, Michalski, Materzynska, Westphal, Kim, Haenel, Fruend, Yianilos, Mueller-Freitag, et~al.]{goyal2017something}
Raghav Goyal, Samira Ebrahimi~Kahou, Vincent Michalski, Joanna Materzynska, Susanne Westphal, Heuna Kim, Valentin Haenel, Ingo Fruend, Peter Yianilos, Moritz Mueller-Freitag, et~al.
\newblock The" something something" video database for learning and evaluating visual common sense.
\newblock In \emph{Proceedings of the IEEE international conference on computer vision}, pages 5842--5850, 2017.

\bibitem[Guzhov et~al.(2022)Guzhov, Raue, Hees, and Dengel]{d-CLIP-guzhov2022audioCLIP}
Andrey Guzhov, Federico Raue, J{\"o}rn Hees, and Andreas Dengel.
\newblock Audioclip: Extending clip to image, text and audio.
\newblock In \emph{ICASSP 2022-2022 IEEE International Conference on Acoustics, Speech and Signal Processing (ICASSP)}, pages 976--980. IEEE, 2022.

\bibitem[He et~al.(2016)He, Zhang, Ren, and Sun]{he2016deep-resnet}
Kaiming He, Xiangyu Zhang, Shaoqing Ren, and Jian Sun.
\newblock Deep residual learning for image recognition.
\newblock In \emph{Proceedings of the IEEE conference on computer vision and pattern recognition}, pages 770--778, 2016.

\bibitem[Kingma and Ba(2014)]{2014Adam}
Diederik Kingma and Jimmy Ba.
\newblock Adam: A method for stochastic optimization.
\newblock \emph{Computer Science}, 2014.

\bibitem[Kuehne et~al.(2011)Kuehne, Jhuang, Garrote, Poggio, and Serre]{kuehne2011hmdb}
Hildegard Kuehne, Hueihan Jhuang, Est{\'\i}baliz Garrote, Tomaso Poggio, and Thomas Serre.
\newblock Hmdb: a large video database for human motion recognition.
\newblock In \emph{2011 International conference on computer vision}, pages 2556--2563. IEEE, 2011.

\bibitem[Kumar~Dwivedi et~al.(2019)Kumar~Dwivedi, Gupta, Mitra, Ahmed, and Jain]{kumar2019protogan}
Sai Kumar~Dwivedi, Vikram Gupta, Rahul Mitra, Shuaib Ahmed, and Arjun Jain.
\newblock Protogan: Towards few shot learning for action recognition.
\newblock In \emph{Proceedings of the IEEE/CVF International Conference on Computer Vision Workshops}, pages 0--0, 2019.

\bibitem[Li et~al.(2022{\natexlab{a}})Li, Zhang, Wu, Jin, and Shan]{li2022hierarchical-HCR}
Changzhen Li, Jie Zhang, Shuzhe Wu, Xin Jin, and Shiguang Shan.
\newblock Hierarchical compositional representations for few-shot action recognition.
\newblock \emph{arXiv preprint arXiv:2208.09424}, 2022{\natexlab{a}}.

\bibitem[Li et~al.(2022{\natexlab{b}})Li, Zhang, Zhang, Yang, Li, Zhong, Wang, Yuan, Zhang, Hwang, et~al.]{d-CLIP-li2022grounded-detect}
Liunian~Harold Li, Pengchuan Zhang, Haotian Zhang, Jianwei Yang, Chunyuan Li, Yiwu Zhong, Lijuan Wang, Lu Yuan, Lei Zhang, Jenq-Neng Hwang, et~al.
\newblock Grounded language-image pre-training.
\newblock In \emph{Proceedings of the IEEE/CVF Conference on Computer Vision and Pattern Recognition}, pages 10965--10975, 2022{\natexlab{b}}.

\bibitem[Li et~al.(2022{\natexlab{c}})Li, Liu, Qian, Li, See, Fei, Yu, and Lin]{li2022ta2n}
Shuyuan Li, Huabin Liu, Rui Qian, Yuxi Li, John See, Mengjuan Fei, Xiaoyuan Yu, and Weiyao Lin.
\newblock Ta2n: Two-stage action alignment network for few-shot action recognition.
\newblock In \emph{Proceedings of the AAAI Conference on Artificial Intelligence}, pages 1404--1411, 2022{\natexlab{c}}.

\bibitem[Liu et~al.(2018)Liu, Lee, Park, Kim, Yang, Hwang, and Yang]{liu2018learning-TNP-few_shot_learning7}
Yanbin Liu, Juho Lee, Minseop Park, Saehoon Kim, Eunho Yang, Sung~Ju Hwang, and Yi Yang.
\newblock Learning to propagate labels: Transductive propagation network for few-shot learning.
\newblock \emph{arXiv preprint arXiv:1805.10002}, 2018.

\bibitem[Lu et~al.(2021)Lu, Ye, and Zhan]{lu2021few-cmot}
Su Lu, Han-Jia Ye, and De-Chuan Zhan.
\newblock Few-shot action recognition with compromised metric via optimal transport.
\newblock \emph{arXiv preprint arXiv:2104.03737}, 2021.

\bibitem[Luo et~al.(2022)Luo, Ji, Zhong, Chen, Lei, Duan, and Li]{luo2022CLIP4CLIP}
Huaishao Luo, Lei Ji, Ming Zhong, Yang Chen, Wen Lei, Nan Duan, and Tianrui Li.
\newblock Clip4clip: An empirical study of clip for end to end video clip retrieval and captioning.
\newblock \emph{Neurocomputing}, 508:\penalty0 293--304, 2022.

\bibitem[Melekhov et~al.(2016)Melekhov, Kannala, and Rahtu]{melekhov2016-siamese—few_shot_learning4}
Iaroslav Melekhov, Juho Kannala, and Esa Rahtu.
\newblock Siamese network features for image matching.
\newblock In \emph{2016 23rd international conference on pattern recognition (ICPR)}, pages 378--383. IEEE, 2016.

\bibitem[M{\"u}ller(2007)]{muller2007dynamic}
Meinard M{\"u}ller.
\newblock Dynamic time warping.
\newblock \emph{Information retrieval for music and motion}, pages 69--84, 2007.

\bibitem[Ni et~al.(2022)Ni, Wen, Liu, Ji, and Yang]{ni2022multi-modal-CLIP}
Xinzhe Ni, Hao Wen, Yong Liu, Yatai Ji, and Yujiu Yang.
\newblock Multimodal prototype-enhanced network for few-shot action recognition.
\newblock \emph{arXiv preprint arXiv:2212.04873}, 2022.

\bibitem[Pahde et~al.(2021)Pahde, Puscas, Klein, and Nabi]{pahde2021multimodal-few-augmentation4}
Frederik Pahde, Mihai Puscas, Tassilo Klein, and Moin Nabi.
\newblock Multimodal prototypical networks for few-shot learning.
\newblock In \emph{Proceedings of the IEEE/CVF Winter Conference on Applications of Computer Vision}, pages 2644--2653, 2021.

\bibitem[Perez and Wang(2017)]{perez2017effectiveness-few-augmentation1}
Luis Perez and Jason Wang.
\newblock The effectiveness of data augmentation in image classification using deep learning.
\newblock \emph{arXiv preprint arXiv:1712.04621}, 2017.

\bibitem[Perrett et~al.(2021)Perrett, Masullo, Burghardt, Mirmehdi, and Damen]{perrett2021temporal-trx}
Toby Perrett, Alessandro Masullo, Tilo Burghardt, Majid Mirmehdi, and Dima Damen.
\newblock Temporal-relational crosstransformers for few-shot action recognition.
\newblock In \emph{Proceedings of the IEEE/CVF Conference on Computer Vision and Pattern Recognition}, pages 475--484, 2021.

\bibitem[Qi et~al.(2020)Qi, Qin, Zhen, Huang, Yang, and Luo]{qi2020few-smfn}
Mengshi Qi, Jie Qin, Xiantong Zhen, Di Huang, Yi Yang, and Jiebo Luo.
\newblock Few-shot ensemble learning for video classification with slowfast memory networks.
\newblock In \emph{Proceedings of the 28th ACM International Conference on Multimedia}, pages 3007--3015, 2020.

\bibitem[Radford et~al.(2021)Radford, Kim, Hallacy, Ramesh, Goh, Agarwal, Sastry, Askell, Mishkin, Clark, et~al.]{CLIP-origin2021}
Alec Radford, Jong~Wook Kim, Chris Hallacy, Aditya Ramesh, Gabriel Goh, Sandhini Agarwal, Girish Sastry, Amanda Askell, Pamela Mishkin, Jack Clark, et~al.
\newblock Learning transferable visual models from natural language supervision.
\newblock In \emph{International conference on machine learning}, pages 8748--8763. PMLR, 2021.

\bibitem[Ratner et~al.(2017)Ratner, Ehrenberg, Hussain, Dunnmon, and R{\'e}]{ratner2017learning-few-augmentation2}
Alexander~J Ratner, Henry Ehrenberg, Zeshan Hussain, Jared Dunnmon, and Christopher R{\'e}.
\newblock Learning to compose domain-specific transformations for data augmentation.
\newblock \emph{Advances in neural information processing systems}, 30, 2017.

\bibitem[Ravi and Larochelle(2017)]{ravi2017optimizationL-few-shot-model4}
Sachin Ravi and Hugo Larochelle.
\newblock Optimization as a model for few-shot learning.
\newblock In \emph{International conference on learning representations}, 2017.

\bibitem[Rusu et~al.(2018)Rusu, Rao, Sygnowski, Vinyals, Pascanu, Osindero, and Hadsell]{rusu2018-metaL-few-shot-model3}
Andrei~A Rusu, Dushyant Rao, Jakub Sygnowski, Oriol Vinyals, Razvan Pascanu, Simon Osindero, and Raia Hadsell.
\newblock Meta-learning with latent embedding optimization.
\newblock \emph{arXiv preprint arXiv:1807.05960}, 2018.

\bibitem[Shao et~al.(2022)Shao, Wu, You, Gao, and Sang]{9999670-generalization-of-MAML}
Yuanjie Shao, Wenxiao Wu, Xinge You, Changxin Gao, and Nong Sang.
\newblock Improving the generalization of maml in few-shot classification via bi-level constraint.
\newblock \emph{IEEE Transactions on Circuits and Systems for Video Technology}, pages 1--1, 2022.

\bibitem[Simon et~al.(2020)Simon, Koniusz, Nock, and Harandi]{simon2020adaptive-few_shot_learning1}
Christian Simon, Piotr Koniusz, Richard Nock, and Mehrtash Harandi.
\newblock Adaptive subspaces for few-shot learning.
\newblock In \emph{Proceedings of the IEEE/CVF conference on computer vision and pattern recognition}, pages 4136--4145, 2020.

\bibitem[Snell et~al.(2017)Snell, Swersky, and Zemel]{snell2017prototypical_few_shot_learning2}
Jake Snell, Kevin Swersky, and Richard Zemel.
\newblock Prototypical networks for few-shot learning.
\newblock \emph{Advances in neural information processing systems}, 30, 2017.

\bibitem[Soomro et~al.(2012)Soomro, Zamir, and Shah]{soomro2012ucf101}
Khurram Soomro, Amir~Roshan Zamir, and Mubarak Shah.
\newblock Ucf101: A dataset of 101 human actions classes from videos in the wild.
\newblock \emph{arXiv preprint arXiv:1212.0402}, 2012.

\bibitem[Su et~al.(2021)Su, Xing, An, Peng, and Feng]{su2021vdarn}
Yong Su, Meng Xing, Simin An, Weilong Peng, and Zhiyong Feng.
\newblock Vdarn: video disentangling attentive relation network for few-shot and zero-shot action recognition.
\newblock \emph{Ad Hoc Networks}, 113:\penalty0 102380, 2021.

\bibitem[Sung et~al.(2018)Sung, Yang, Zhang, Xiang, Torr, and Hospedales]{sung2018learning-relation-networkfew_shot_learning3}
Flood Sung, Yongxin Yang, Li Zhang, Tao Xiang, Philip~HS Torr, and Timothy~M Hospedales.
\newblock Learning to compare: Relation network for few-shot learning.
\newblock In \emph{Proceedings of the IEEE conference on computer vision and pattern recognition}, pages 1199--1208, 2018.

\bibitem[Tan and Yang(2019)]{tan2019learning-FAN}
Shaoqing Tan and Ruoyu Yang.
\newblock Learning similarity: Feature-aligning network for few-shot action recognition.
\newblock In \emph{2019 International Joint Conference on Neural Networks (IJCNN)}, pages 1--7. IEEE, 2019.

\bibitem[Thatipelli et~al.(2022)Thatipelli, Narayan, Khan, Anwer, Khan, and Ghanem]{thatipelli2022spatio-strm}
Anirudh Thatipelli, Sanath Narayan, Salman Khan, Rao~Muhammad Anwer, Fahad~Shahbaz Khan, and Bernard Ghanem.
\newblock Spatio-temporal relation modeling for few-shot action recognition.
\newblock In \emph{Proceedings of the IEEE/CVF Conference on Computer Vision and Pattern Recognition}, pages 19958--19967, 2022.

\bibitem[Vaswani et~al.(2017)Vaswani, Shazeer, Parmar, Uszkoreit, Jones, Gomez, Kaiser, and Polosukhin]{vaswani2017attention}
Ashish Vaswani, Noam Shazeer, Niki Parmar, Jakob Uszkoreit, Llion Jones, Aidan~N Gomez, {\L}ukasz Kaiser, and Illia Polosukhin.
\newblock Attention is all you need.
\newblock \emph{Advances in neural information processing systems}, 30, 2017.

\bibitem[Vinyals et~al.(2016)Vinyals, Blundell, Lillicrap, Wierstra, et~al.]{vinyals2016matching—few_shot_learning5}
Oriol Vinyals, Charles Blundell, Timothy Lillicrap, Daan Wierstra, et~al.
\newblock Matching networks for one shot learning.
\newblock \emph{Advances in neural information processing systems}, 29, 2016.

\bibitem[Wang et~al.(2022{\natexlab{a}})Wang, Jin, Feng, and Li]{wang2022task}
Jiayi Wang, Yi Jin, Songhe Feng, and Yidong Li.
\newblock Task adaptive modeling for few-shot action recognition.
\newblock In \emph{2022 IEEE 24th International Workshop on Multimedia Signal Processing (MMSP)}, pages 1--6. IEEE, 2022{\natexlab{a}}.

\bibitem[Wang et~al.(2016)Wang, Xiong, Wang, Qiao, Lin, Tang, and Van~Gool]{wang2016temporal-tsn}
Limin Wang, Yuanjun Xiong, Zhe Wang, Yu Qiao, Dahua Lin, Xiaoou Tang, and Luc Van~Gool.
\newblock Temporal segment networks: Towards good practices for deep action recognition.
\newblock In \emph{European conference on computer vision}, pages 20--36. Springer, 2016.

\bibitem[Wang et~al.(2021{\natexlab{a}})Wang, Xing, and Liu]{wang2021actionCLIP}
Mengmeng Wang, Jiazheng Xing, and Yong Liu.
\newblock Actionclip: A new paradigm for video action recognition.
\newblock \emph{arXiv preprint arXiv:2109.08472}, 2021{\natexlab{a}}.

\bibitem[Wang et~al.(2021{\natexlab{b}})Wang, Ye, Qi, Zhao, Wang, Shan, and Wang]{wang2021semantic}
Xiao Wang, Weirong Ye, Zhongang Qi, Xun Zhao, Guangge Wang, Ying Shan, and Hanzi Wang.
\newblock Semantic-guided relation propagation network for few-shot action recognition.
\newblock In \emph{Proceedings of the 29th ACM International Conference on Multimedia}, pages 816--825, 2021{\natexlab{b}}.

\bibitem[Wang et~al.(2022{\natexlab{b}})Wang, Zhang, Qing, Tang, Zuo, Gao, Jin, and Sang]{wang2022-hybrid}
Xiang Wang, Shiwei Zhang, Zhiwu Qing, Mingqian Tang, Zhengrong Zuo, Changxin Gao, Rong Jin, and Nong Sang.
\newblock Hybrid relation guided set matching for few-shot action recognition.
\newblock In \emph{Proceedings of the IEEE/CVF Conference on Computer Vision and Pattern Recognition}, pages 19948--19957, 2022{\natexlab{b}}.

\bibitem[Wang et~al.(2023{\natexlab{a}})Wang, Ye, Qi, Wang, Wu, Shan, Qie, and Wang]{2023Task-AwareDual-Representation}
Xiao Wang, Weirong Ye, Zhongang Qi, Guangge Wang, Jianping Wu, Ying Shan, Xiaohu Qie, and Hanzi Wang.
\newblock Task-aware dual-representation network for few-shot action recognition.
\newblock \emph{IEEE Transactions on Circuits and Systems for Video Technology}, pages 1--1, 2023{\natexlab{a}}.

\bibitem[Wang et~al.(2023{\natexlab{b}})Wang, Zhang, Cen, Gao, Zhang, Zhao, and Sang]{wang2023CLIP}
Xiang Wang, Shiwei Zhang, Jun Cen, Changxin Gao, Yingya Zhang, Deli Zhao, and Nong Sang.
\newblock Clip-guided prototype modulating for few-shot action recognition.
\newblock \emph{arXiv preprint arXiv:2303.02982}, 2023{\natexlab{b}}.

\bibitem[Wang et~al.(2023{\natexlab{c}})Wang, Zhang, Qing, Gao, Zhang, Zhao, and Sang]{wang2023molo}
Xiang Wang, Shiwei Zhang, Zhiwu Qing, Changxin Gao, Yingya Zhang, Deli Zhao, and Nong Sang.
\newblock Molo: Motion-augmented long-short contrastive learning for few-shot action recognition.
\newblock \emph{arXiv preprint arXiv:2304.00946}, 2023{\natexlab{c}}.

\bibitem[Wanyan et~al.(2023)Wanyan, Yang, Chen, and Xu]{wanyan2023active}
Yuyang Wanyan, Xiaoshan Yang, Chaofan Chen, and Changsheng Xu.
\newblock Active exploration of multimodal complementarity for few-shot action recognition.
\newblock In \emph{Proceedings of the IEEE/CVF Conference on Computer Vision and Pattern Recognition}, pages 6492--6502, 2023.

\bibitem[Wu et~al.(2022)Wu, Zhang, Zhang, Wu, and Zhang]{wu2022motion}
Jiamin Wu, Tianzhu Zhang, Zhe Zhang, Feng Wu, and Yongdong Zhang.
\newblock Motion-modulated temporal fragment alignment network for few-shot action recognition.
\newblock In \emph{Proceedings of the IEEE/CVF Conference on Computer Vision and Pattern Recognition}, pages 9151--9160, 2022.

\bibitem[Xu et~al.(2022)Xu, De~Mello, Liu, Byeon, Breuel, Kautz, and Wang]{d-CLIP-xu2022groupvit}
Jiarui Xu, Shalini De~Mello, Sifei Liu, Wonmin Byeon, Thomas Breuel, Jan Kautz, and Xiaolong Wang.
\newblock Groupvit: Semantic segmentation emerges from text supervision.
\newblock In \emph{Proceedings of the IEEE/CVF Conference on Computer Vision and Pattern Recognition}, pages 18134--18144, 2022.

\bibitem[Ye et~al.(2020)Ye, Hu, Zhan, and Sha]{ye2020few-embedding-adaptation-few_shot_learning8}
Han-Jia Ye, Hexiang Hu, De-Chuan Zhan, and Fei Sha.
\newblock Few-shot learning via embedding adaptation with set-to-set functions.
\newblock In \emph{Proceedings of the IEEE/CVF conference on computer vision and pattern recognition}, pages 8808--8817, 2020.

\bibitem[Yoon et~al.(2019)Yoon, Seo, and Moon]{yoon2019tapnet}
Sung~Whan Yoon, Jun Seo, and Jaekyun Moon.
\newblock Tapnet: Neural network augmented with task-adaptive projection for few-shot learning.
\newblock In \emph{International conference on machine learning}, pages 7115--7123. PMLR, 2019.

\bibitem[Zhang et~al.(2020{\natexlab{a}})Zhang, Zhang, Qi, Li, Torr, and Koniusz]{zhang2020few-ARN}
Hongguang Zhang, Li Zhang, Xiaojuan Qi, Hongdong Li, Philip~HS Torr, and Piotr Koniusz.
\newblock Few-shot action recognition with permutation-invariant attention.
\newblock In \emph{Computer Vision--ECCV 2020: 16th European Conference, Glasgow, UK, August 23--28, 2020, Proceedings, Part V 16}, pages 525--542. Springer, 2020{\natexlab{a}}.

\bibitem[Zhang et~al.(2020{\natexlab{b}})Zhang, Chang, Liu, Luo, Prakash, and Hauptmann]{zhang2020few1}
Lingling Zhang, Xiaojun Chang, Jun Liu, Minnan Luo, Mahesh Prakash, and Alexander~G Hauptmann.
\newblock Few-shot activity recognition with cross-modal memory network.
\newblock \emph{Pattern Recognition}, 108:\penalty0 107348, 2020{\natexlab{b}}.

\bibitem[Zhang et~al.(2022)Zhang, Zeng, Guo, and Li]{d-CLIP-zhang2022can-depth}
Renrui Zhang, Ziyao Zeng, Ziyu Guo, and Yafeng Li.
\newblock Can language understand depth?
\newblock In \emph{Proceedings of the 30th ACM International Conference on Multimedia}, pages 6868--6874, 2022.

\bibitem[Zheng et~al.(2022)Zheng, Chen, and Jin]{zheng2022few-hcl}
Sipeng Zheng, Shizhe Chen, and Qin Jin.
\newblock Few-shot action recognition with hierarchical matching and contrastive learning.
\newblock In \emph{European Conference on Computer Vision}, pages 297--313. Springer, 2022.

\bibitem[Zhu and Yang(2018)]{zhu2018compound}
Linchao Zhu and Yi Yang.
\newblock Compound memory networks for few-shot video classification.
\newblock In \emph{Proceedings of the European Conference on Computer Vision (ECCV)}, pages 751--766, 2018.

\bibitem[Zhu and Yang(2020)]{zhu2020label}
Linchao Zhu and Yi Yang.
\newblock Label independent memory for semi-supervised few-shot video classification.
\newblock \emph{IEEE Transactions on Pattern Analysis and Machine Intelligence}, 44\penalty0 (1):\penalty0 273--285, 2020.

\end{thebibliography}
}


\end{document}